\documentclass[10pt,twocolumn,letterpaper]{article}

\usepackage{cvpr}      %

\usepackage[dvipsnames,table]{xcolor}

\definecolor{amethyst}{rgb}{0.6, 0.4, 0.8}
\definecolor{darkpastelgreen}{rgb}{0.01, 0.75, 0.24}
\definecolor{amber}{rgb}{1.0, 0.75, 0.0}
\definecolor{cadmiumorange}{rgb}{0.93, 0.53, 0.18}
\definecolor{lawngreen}{rgb}{0.49, 0.99, 0.0}
\definecolor{limegreen}{rgb}{0.2, 0.8, 0.2}
\definecolor{neongreen}{rgb}{0.22, 0.88, 0.08}
\definecolor{amethyst}{rgb}{0.6, 0.4, 0.8}
\definecolor{darkpastelgreen}{rgb}{0.01, 0.75, 0.24}
\definecolor{greenbest}{RGB}{88,137,15}
\definecolor{redworst}{RGB}{137,15,27}
\definecolor{royalazure}{rgb}{0.25, 0.41, 0.88}
\usepackage{multirow}
\usepackage{stix}
\usepackage{colortbl}
\newcommand\sbullet[1][.5]{\mathbin{\vcenter{\hbox{\scalebox{#1}{$\bullet$}}}}}

\definecolor{tabfirst}{rgb}{1, 0.7, 0.7} 
\definecolor{tabsecond}{rgb}{1, 0.85, 0.7} 
\definecolor{tabthird}{rgb}{1, 1, 0.7} 

\usepackage{pifont}

\newcommand{\Plus}{\mathord{\text{\ding{58}}}}

\usepackage{soul}

\usepackage{amsmath}
\DeclareMathOperator*{\argmax}{arg\,max}

\usepackage{comment}

\definecolor{cvprblue}{rgb}{0.21,0.49,0.74}
\usepackage[pagebackref,breaklinks,colorlinks,citecolor=cvprblue]{hyperref}
\usepackage[accsupp]{axessibility}

\def\paperID{6348} %
\def\confName{CVPR}
\def\confYear{2024}

\begin{document}

\title{TexTile: A Differentiable Metric for Texture Tileability}

\author{Carlos Rodriguez-Pardo$^{1}$~~~~~~~~~~Dan Casas$^{1}$~~~~~~~~~~Elena Garces$^{1,2}$~~~~~~~~~~Jorge Lopez-Moreno$^{1,2}$\\[0.2cm]
 $^1$Universidad Rey Juan Carlos, Spain ~~ $^2$SEDDI, Spain  \\[0.1cm]
}

\twocolumn[{%
	\renewcommand\twocolumn[1][]{#1}%
	\maketitle
	\begin{center}
		\vspace{-0.65cm}
			\includegraphics[width=1.0\linewidth]{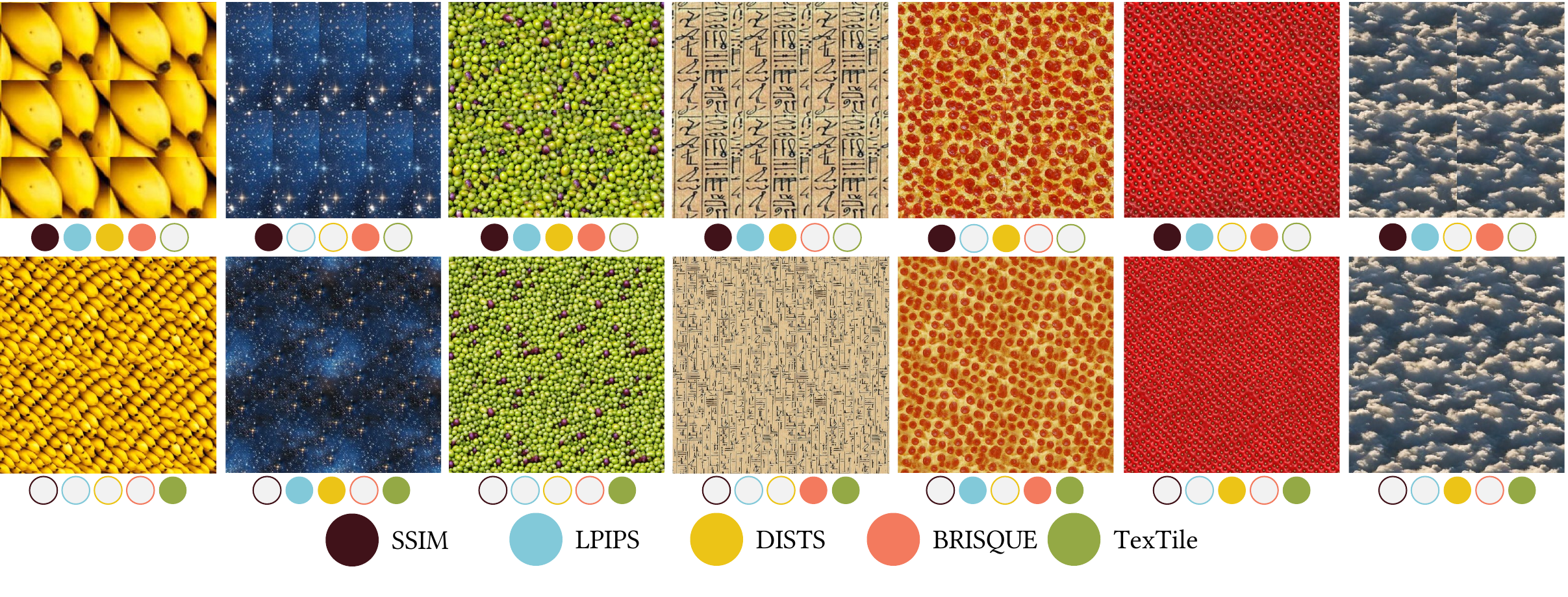}
		\vspace{-0.85cm}
		\captionof{figure}{Existing perceptual metrics, commonly used to evaluate texture synthesis algorithms, typically fail to account for tileability.
  Such weakness is depicted in this figure where, for each column, we show tiled versions of textures with (top) and without (bottom) tiling artifacts.
  For each column, we highlight using saturated color dots the preferred image (\textit{i.e.}, higher score) according to different metrics.
  It can be seen that there is no correlation across existing methods (\textit{i.e.}, saturated dots distributed over top and bottom rows), while our method TexTile consistently prefers seamless tiled textures (\textit{i.e.,} saturated green dots on the bottom for all columns).}
	\label{fig:teaser}
		\vspace{-0.2cm}
	\end{center}
}]

\thispagestyle{empty}

\maketitle

\begin{abstract}
We introduce TexTile, a novel differentiable metric to quantify the degree upon which a texture image can be concatenated with itself without introducing repeating artifacts (i.e., the tileability). Existing methods for tileable texture synthesis focus on general texture quality, but lack explicit analysis of the intrinsic repeatability properties of a texture. In contrast, our TexTile metric effectively evaluates the tileable properties of a texture, opening the door to more informed synthesis and analysis of tileable textures. Under the hood, TexTile is formulated as a binary classifier carefully built from a large dataset of textures of different styles, semantics, regularities, and human annotations. Key to our method is a set of architectural modifications to baseline pre-train image classifiers to overcome their shortcomings at measuring tileability, along with a custom data augmentation and training regime aimed at increasing robustness and accuracy. We demonstrate that TexTile can be plugged into different state-of-the-art texture synthesis methods, including diffusion-based strategies, and generate tileable textures while keeping or even improving the overall texture quality. Furthermore, we show that TexTile can objectively evaluate any tileable texture synthesis method, whereas the current mix of existing metrics produces uncorrelated scores which heavily hinders progress in the field.

\end{abstract} \vspace{-5mm}

\section{Introduction}
\label{sec:introduction}
The appearance of 3D digital objects plays a fundamental role in the overall realism of a virtual environment.
To create realistic textures, many strategies have been widely explored, including procedural algorithms \cite{GLLD2012GNBE,hu2019novel}, scanning \cite{guo2020materialgan,moritz2017texture} and, more recently, text-to-image generative pipelines \cite{chen2023text2tex,cao2023texfusion,casas2023smplitex}. 
Among the different properties that we wish for the synthesized textures (\textit{e.g.}, photorealism, variety in detail, high resolution), the ability to seamlessly repeat or tile itself without noticeable artifacts ––its tileability–– is especially important in the frequent case of applying a texture to a large surface.
For example, when texturing the facade of a building or a field of grass. 

Many methods exist that focus on the specific case of tileable texture synthesis~\cite{niklasson2021self-organising,moritz2017texture,li2020inverse,deliot2019procedural,rodriguez2019automatic,aigerman2023generative, rodriguezseamlessgan}.
This has been achieved, for example, by manipulating image borders \cite{li2020inverse}, maximizing stationary image  properties \cite{moritz2017texture}, or conditioning generative models on structured patterns \cite{zhou2022tilegen}.
However, despite such significantly diverse methodologies used in existing methods, they typically rely on evaluation using common metrics based on general texture quality which, unfortunately, do not explicitly account for the intrinsic repeatability properties of a texture.  

To address this shortcoming, we introduce TexTile, a novel metric for texture tileability.
TexTile is a data-driven metric that brings two key novel functionalities into texture synthesis methods:
first, it computes a human-friendly score that captures the intrinsic repeatability of any texture; 
and second, it provides a differentiable metric that can be used as an additional data term in any learning-based or test-time optimization method for tileable texture synthesis.
We demonstrate that TexTile enables a factual analysis of state-of-the-art methods for tileable texture synthesis, while previous metrics often result in uncorrelated evaluations (\textit{i.e.}, a good tileable texture might have low SSIM~\cite{wang2004image} score, or a poor tileable texture might have a high SSIM), as illustrated in Figure~\ref{fig:teaser}.
Furthermore, we demonstrate that TexTile can be used off-the-shelf as an additional loss term in state-of-the-art methods for texture synthesis, including diffusion-based models, to output tileable textures while preserving or even improving the overall image quality.

Under the hood, we formulate TexTile as a binary classifier built using a carefully designed architecture with an attention-enhanced convolutional network.
The convolutional filters can detect local discontinuities --which are common in borders that are not seamlessly tileable-- and can deal with images of arbitrary sizes, but they struggle with global understanding to detect artifacts and repeating patterns.
Therefore, we introduce Self-Attention layers into our architecture, which is known to capture a global understanding of the input.
This, combined with a custom data augmentation policy designed for tileability detection, enables us to train a novel neural classifier to unleash a new functionality for the state-of-the-art texture synthesis methods.

In summary, we introduce the following contributions:
\begin{itemize}
    \item A novel learning-based metric for texture analysis that accurately quantifies tileability.
    \item An attention-enhanced convolutional classifier, and a training configuration aimed at maximizing robustness and accuracy.
    \item A differentiable loss function which can be plugged into texture synthesis algorithms to generate tileable textures. 
    \item Open-source code and trained weights for our metric. We believe this will open the door to quantitative benchmarks on tileable textures, which is currently not possible due to the lack of a specific metric for such task.
\end{itemize}

\section{Related Work}\label{sec:related_work}
\begin{figure}[tb!]
	\centering
	\includegraphics[width=1.0\columnwidth]{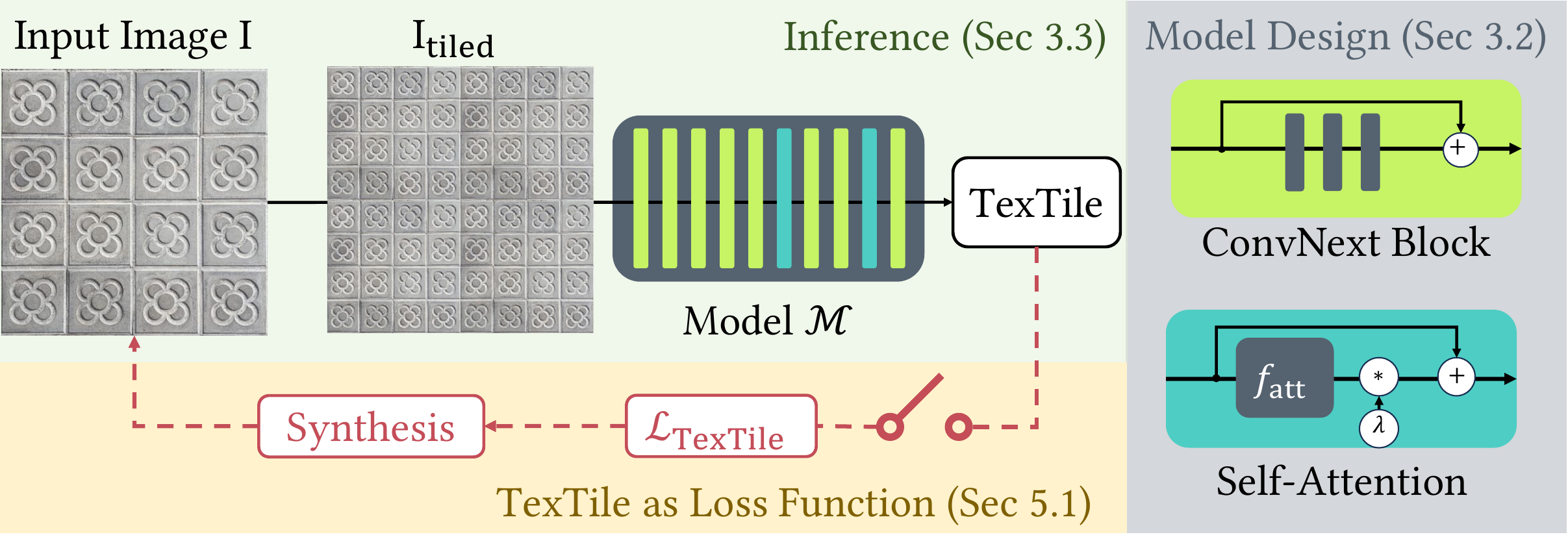}
	\vspace{-7mm}
	\caption{Our model takes as input a texture image $\text{I}$, which we tile to form $\text{I}_{\text{tiled}}$, and returns an estimation of its tileability. This metric can be used as a loss function $\mathcal{L}_{\text{TexTile}}$ to allow synthesis algorithms to generate tileable textures. Our model, $\mathcal{M}$ architecture is comprised of ConvNext~\cite{liu2022convnet} and residual self-attention blocks.}
	\label{fig:textile}
 \vspace{-5mm}
\end{figure}
\subsection{Image Quality Assessment}

\label{sec:iqa}
Image Quality Assessment (IQA) algorithms can be categorized into three different groups. 
\textbf{Reference-based} IQA methods compare an input and a reference image, which is the most widely studied strategy for IQA.
These methods have traditionally leveraged pixel-wise differences (\textit{e.g.,} PSNR, $\ell_1$ or $\ell_2$ distances) or image statistics (\textit{e.g.}, SSIM~\cite{wang2004image} or FSIM~\cite{zhang2011fsim}) to compute the similarity between the images.
Neural reference-based IQA, which leverage the stronger correlation of deep neural networks with human perception~\cite{zhang2018unreasonable}, have been also proposed.
These strategies either leverage features from untrained~\cite{amir2021understanding} or pre-trained convolutional neural networks~\cite{gatys2015neural}, or use direct supervision from human judgments, as in LPIPS~\cite{zhang2018unreasonable}, PIE-APP~\cite{prashnani2018pieapp}, DISTS~\cite{ding2020image}, Si-FID~\cite{rottshaham2019singan}, or DreamSIM~\cite{fu2023dreamsim}.
These methods introduce powerful and differentiable metrics, however, they require a reference image and are thus not suitable for measuring global properties, like tileability.

Instead of comparing an input and a reference image, \textbf{distribution-based} IQA methods compare statistics of two sets of reference images and generated images.
These methods are commonly used to evaluate the perceptual quality of generative models, typically using metrics based on neural networks activations \cite{salimans2016improved,heusel2017gans,binkowski2018demystifying,xie2023learning}, nearest neighbors~\cite{kynkaanniemi2019improved}, spectral~\cite{tsitsulin2019shape}, or geometric distances~\cite{khrulkov2018geometry}.
While these methods are useful for evaluating the performance of generative models, they struggle as loss functions~\cite{salimans2016improved}, and also require reference images.

Finally, \textbf{no-reference} IQA methods compute the overall quality score of an input image without requiring an explicit reference.
These methods rely on image statistics, as in BIQI~\cite{moorthy2010two} amd BRISQUE~\cite{mittal2012no}; or on training deep neural networks on human judgments of image quality, like HyperIQA~\cite{su2020blindly}, MANIQA~\cite{yang2022maniqa}, VCRNet~\cite{pan2022vcrnet}, or CLIP-IQA~\cite{wang2023exploring}.
Despite competitive results in image quality assessment, these methods do not incorporate tileabilty analysis. 
SeamlessGAN~\cite{rodriguezseamlessgan} leverages the discriminator of a single-image generative model to find artifacts in the borders of generated textures. However, the discriminator only measures seamlessness, ignoring other factors that influence tileability and, most importantly, it cannot generalize to any image outside of its single-image dataset.

Our metric is most closely related to no-reference IQA methods, as it takes a single image as input.
However, in contrast to existing methods, we assess the image quality based on tilebility instead of general image quality. %
We demonstrate that, when combined with existing IQA metrics, our metric successfully captures overall quality \textit{and} tileability. %
To the best of our knowledge, our metric is the first no-reference tileability metric.

\subsection{Tileable Texture Synthesis}

\textbf{Non-parametric} tileable texture synthesis methods generate new tileable images by maximizing image~\emph{stationarity}~\cite{moritz2017texture}, by manipulating the images with border transformations using Graphcuts~\cite{li2020inverse}, or by patch-based synthesis with histogram-preserving blending~\cite{deliot2019procedural}. \textbf{Parametric} alternatives typically leverage deep neural networks in diverse ways.
Rodriguez-Pardo \etal~\cite{rodriguez2019automatic} look for repeating patterns in images using deep features in pretrained CNNs, then synthesize tileable images by blending the borders.
Tileability can be also achieved with specific image parameterizations and neural network design, as in Neural Cellular Automata~\cite{niklasson2021self-organising}, or in the Periodic Spatial GAN~\cite{bergmann2017learning}.
By manipulating latent spaces in pre-trained GANs, SeamlessGAN~\cite{rodriguezseamlessgan} achieve tileability without specific model modifications. 
Tileable image generation has also been explored with the goal of material capture, by means of models conditioned on structured patterns~\cite{zhou2022tilegen}, or through \emph{rolled diffusion}~\cite{vecchio2023controlmat}. 
Specialized methods have been proposed for tileable vector image generation~\cite{aigerman2023generative}.
For a review on texture synthesis, we refer the reader to the survey in~\cite{akl2018survey}. 

These methods typically provide quantitative evaluation using reference-based IQA.
As mentioned in Section \ref{sec:iqa}, these metrics measure the perceptual similarity between generated and input images, however, they do not account for tiling. 
To the best of our knowledge, there is no available metric that can be used to compare these methods in terms of tileability, hindering the progress of the field and limiting evaluation to qualitative analyses. %
Our metric aims to solve this gap. Being fully differentiable, it can be leveraged as a loss function to enable existing synthesis algorithms to produce seamlessly tileable outputs.

\section{TexTile}\label{sec:textile}

\begin{figure}[tb!]
	\centering
		\vspace{-3mm}
	\includegraphics[width=1.0\columnwidth]{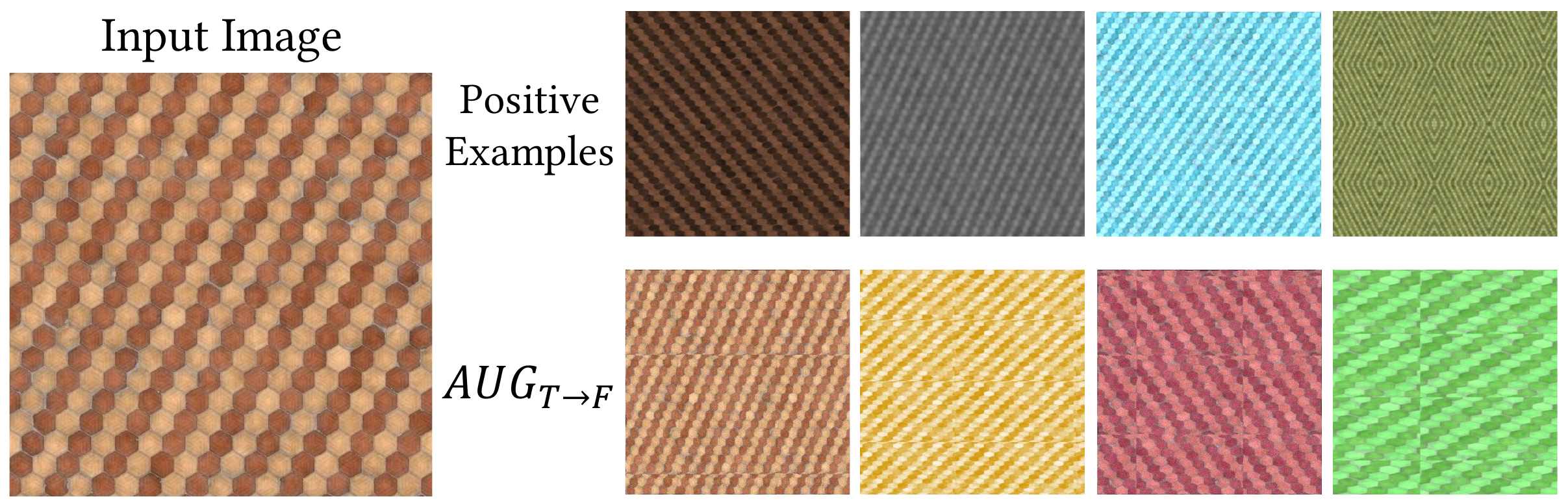}
	\vspace{-5mm}
	\caption{From a tileable texture (left), our data augmentation can generate tileable (top row) and non-tileable (bottom) variations.  }
	\label{fig:data_aug}
		\vspace{-5mm}
\end{figure}

\subsection{Introduction}
Our goal is to develop a differentiable no-reference image metric that measures texture tileability, that is, a single-image metric that does not require a second image for comparison purposes. To achieve this, we leverage a convolutional neural network as our differentiable function, which we train on a dataset of textures on a binary classification task. Our model learns to classify between tileable and non-tileable textures, by means of a comprehensive data augmentation policy and custom architecture design choices.
We explain our model design choices in Sec.~\ref{sec:design}, validate them using ablation studies in Sec.~\ref{sec:ablation}, and explain the model predictions in Sec.~\ref{sec:qualitative}. We show examples of results of TexTile as a loss function (Sec.~\ref{sec:loss}), as a means of benchmarking image synthesis algorithms (Sec.~\ref{sec:benchmark}), and applications for alignment and repeating pattern detection (Sec.~\ref{sec:other_applications}).

 \subsection{Model Design and Training}\label{sec:design}

\subsubsection*{Network Design}\label{sec:network}
A model that can measure texture tileability should have at least three properties. First, it must be able to detect local discontinuities, which happen when borders are not seamlessly tileable. Second, the model should be able to handle images of any dimension or aspect ratios. Finally, it should have a global understanding of the image, in order to detect artifacts and repeating patterns. The first two properties can be achieved by fully-convolutional architectures, which are strongly biased towards textures~\cite{hermann2020origins}. However, problems that require global understanding of images are typically tackled using attention-based Vision Transformers~\cite{han2022survey}, which are more biased towards shape~\cite{naseer2021intriguing} and are less flexible in terms of input dimensionality.

We propose an architecture that can benefit from the properties of both convolutional and attention-based models. Because of the limited size of our training dataset, we also want to leverage ImageNet pretraining~\cite{deng2009imagenet}. We thus use a state-of-the-art pretrained ConvNext~\cite{liu2022convnet} fully-convolutional model. We further introduce Linear Self-Attention modules~\cite{wang2020linformer} to allow it to learn global patterns in the images, while keeping a limited computational cost. We design them as residual layers $x \gets x + \lambda f_{\text{att}}(x)$, multiplied by a learnable parameter $\lambda$, which we initialize at $1e-6$. We illustrate this module in Figure~\ref{fig:textile}. This way, we can modify the internal model architecture without disrupting the performance of the pre-trained backbone during early training iterations. We only use two of such modules, which we place on the deeper layers of the ConvNext model. Previous work on texture estimation and synthesis also benefit from adding attention to fully-convolutional backbones~\cite{guo2022u, rodriguezpardo2023UMat}.

\subsubsection*{Data Augmentation for Tileable Textures}\label{sec:data_aug}

We leverage a comprehensive data augmentation policy, to train the model to detect repeating artifacts in images. 

Our policy contains several operations which we divide into: \textbf{Tileability-preserving} policies, which generate variations of textures without reducing their tileability. These include global color and gamma changes, random flipping (horizontal and vertical), translations, equalization, blurs or noise, or rescaling the images with different scale factors across each dimensions.  We also introduce an operation named \emph{UnFold}, which mirrors the input image horizontally and vertically. \textbf{Tileability-breaking} policies include rotations, shears, random cropping, or thin-plate spline warping~\cite{bartoli2010generalized}. We apply tileability-preserving policies to tileable examples, and both kinds of operations to non-tileable examples. Finally, our last operation is random tiling, by which we tile the textures a random number of times (1-5), followed by a random rescaling, and random cropping. With this policy, we allow the model to detect tileability regardless on the number of repetitions in the input images.

We introduce two additional data augmentation policies. With $AUG_{T\to F}$, we generate non-tileable textures from tileable textures. We do this by applying a tileability-breaking operation to a tileable texture. With $AUG_{F\to F'}$, we create repetition-free texture examples by applying every data augmentation operation except random tiling to non-tileable examples, and assign these textures a positive tileability label. These two policies are needed to reduce the distribution shift between tileable and non-tileable training datasets, as they come from different sources and their semantics, feature scales, and contents may differ. With $AUG_{T\to F}$ and $AUG_{F\to F'}$, we force the model to learn patterns exclusively related to tileability, ignoring other texture properties. We illustrate some of these policies in Figure~\ref{fig:data_aug}, including example of \emph{UnFolding} on the top right image.

\subsection{Model Inference}\label{sec:inference}

We illustrate the inference process in Figure~\ref{fig:textile}. We tile the input image $\text{I}$ once across each spatial dimension $\text{I}_{\text{tiled}} \gets tile(\text{I}, (2,2))$. We observe that this limited number of repetitions is enough to accurately detect tileability.  We use this process both in texture evaluation and when we use TexTile as loss function.  $\text{I}_{\text{tiled}}$ is fed to our model $\mathcal{M}$, which outputs an unbounded prediction, which can be transformed into the desired $(0,1)$ range with a Logistic function. However, doing this typically leads to predictions very close to 0 or 1, making our metric less useful for comparing different textures. We mitigate this issue by introducing $\lambda$, which controls how far the predicted values are from the boundaries: 

\begin{equation}
	\text{TexTile} = \frac{1}{{1 + \exp\left(-\lambda \cdot \mathcal{M}(\text{I}_{\text{tiled}})\right)}}
\end{equation}

We set $\lambda=0.25$, which preserves discrimination between clearly tileable and non-tileable samples while ambiguous cases sit closer to $0.5$. This is a strictly monotonic function, so relative tileability orders are maintained.

\subsection{Dataset}\label{sec:dataset}
 Our goal is to make our model robust to textures of any semantics, regularity, stochasticity, and homogeneity. We also want our model to work with both natural and synthetic texture maps. To achieve this, we collect a dataset of tileable and non-tileable textures from a variety of sources.

We gather a novel dataset of $4276$ tileable textures, comprised of high-resolution photographs of materials, turned into tileable textures by artist labor. We extend this data with publicly-available tileable textures, including $285$ images from~\cite{Vidanapathirana2021Plan2Scene}, $186$ \emph{CC-BY} images from \emph{\href{https://juliosillet.gumroad.com/}{Julio Sillet}}, including both albedo and surface normal maps, and $290$ royalty-free textures from \emph{\href{https://manytextures.com/}{ManyTextures}}. These images, being curated by human experts, ensure the \emph{tileability} property.%

\begin{figure}[tb!]
	\centering
		\vspace{-2mm}
	\includegraphics[width=1.0\columnwidth]{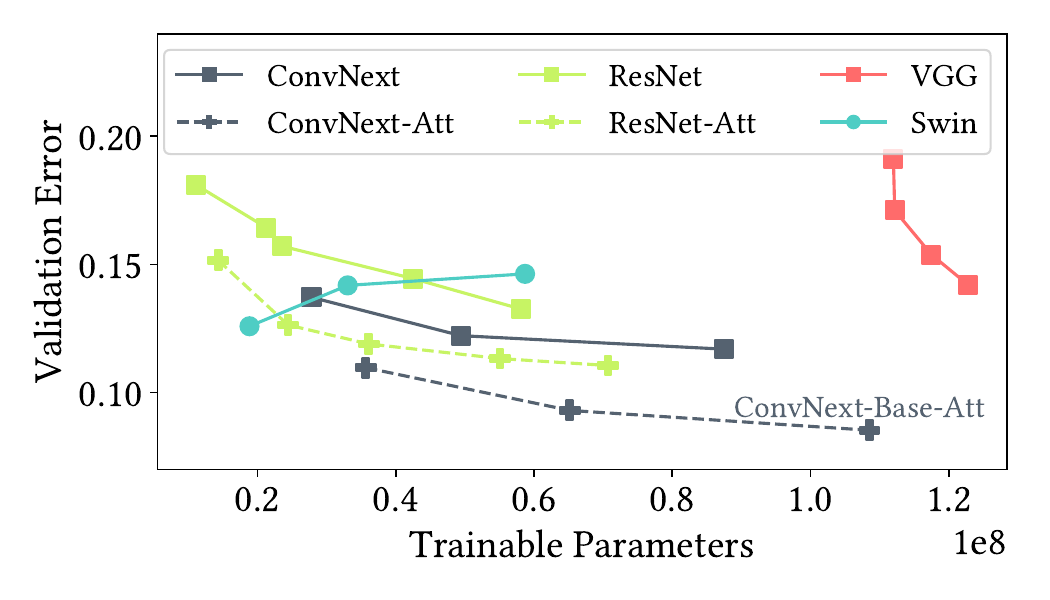}
	\vspace{-8mm}
	\caption{Influence on the neural architecture type and size on its quantitative performance (Cross-entropy error on the validation dataset). Convolutional architectures are marked with ${ \scriptstyle \blacksquare }$, attention-based models with $\scriptstyle \sbullet[2.]$, and our versions of convolutional networks with embedded attention with ${ \footnotesize \Plus }$.  }
	\label{fig:architectures_ablation}
	\vspace{-5mm}
\end{figure}

We also create an additional dataset of non-tileable textures, for which examples are comparably easier to obtain. To this end, we gather $3922$ synthetic high-quality from~\cite{deschaintre2018single}, $1187$ photographs of a wide variety of texture types from~\cite{cimpoi14describing}, $506$ facade photographs from~\cite{isola2017image}, $7265$ images taken under different illumination conditions from~\cite{xilong2021ASSE}, $760$ from~\cite{Zhou2022look-ahead}, and $93$ from~\cite{moritz2017texture}. We extend this dataset with $8675$ images from~\cite{Fritz2004THEKD, Mallikarjuna2006THEK2, Kylberg2011c,lazebnik:inria-00548530}, and $86$ non-stationary textures from~\cite{zhou2018non}. While we can create negative examples from tileable textures using $AUG_{T\to F}$, this dataset is important for  generalization.
We use $504$ tileable images and $504$ non-tileable examples for validation, and the rest is left for training. We have approximately 5 times more non-tileable examples than tileable images, which may hinder the task of learning an unbiased classifier.  We overcome this with our training and data augmentation policies. %

In Section~\ref{sec:evaluation} we evaluate the performance of our model, measuring average binary cross-entropy error on our test set, as well as classification metrics, like accuracy, $F_1$-Score and Area Under the Curve (AUC). We provide more details of our datasets in the supplementary material.

\subsection{Implementation Details}
We train our models for 100 epochs using NAdam~\cite{dozat2016incorporating} Lookahead ~\cite{zhang2019lookahead}, Automatic Gradient Scaling and Mixed Precision Training~\cite{micikevicius2018mixed}, with an initial learning rate of $0.002$, halved every 33 epochs. We use PyTorch ~\cite{paszke2017automatic} for training, and Kornia~\cite{riba2020kornia} for data augmentation. We use batch sizes of 24 samples, composed of balanced number of positive and negative examples. This process takes approximately 6 hours on a single Nvidia RTX 3060 GPU. We use images of $(384, 384)$ pixels for training and $(512,512)$ for inference. Hyperparameters are tuned using Bayesian optimization on a validation dataset containing 1000 images.

\section{Evaluation}\label{sec:evaluation}
\subsection{Ablation Study}\label{sec:ablation}

\paragraph{Network Architecture Design}
In Figure~\ref{fig:architectures_ablation}, we show the impact of the neural network architecture design on generalization performance. We evaluate different fully-convolutional backbones, including ResNet~\cite{he2016deep}, VGG~\cite{simonyan2015very} and ConvNext~\cite{liu2022convnet}; as well as transformer-based Swin V2~\cite{liu2022swin}, on different model sizes. We also show the results of our variations of the fully-convolutional backbones, where we introduce linear Self-Attention~\cite{wang2020linformer} modules in the last layers of the models. Every network is pre-trained on ImageNet~\cite{deng2009imagenet}, then fine tuned in our task, and evaluated on a validation dataset. As shown, larger networks typically perform better, with the exception of Swin, which interestingly benefits from fewer parameters. Further, ConvNexts strongly outperform ResNets and VGGs. By introducing self-attention into these fully-convolutional models, we significantly improve their generalization capabilities. In the rest of our experiments, we will use our custom \emph{ConvNext-Base-Att} model, as it achieves the lowest error overall.

\begin{table}[]
	\centering
        \resizebox{\columnwidth}{!}{
	\begin{tabular}{ c l    @{\extracolsep{\fill}}   cccc}
         \cline{2-6} 
		& \multicolumn{1}{c}{\textbf{Configuration}} & Error  $\downarrow$ & Accuracy $\uparrow$ &  $F_1$ $\uparrow$ & AUC $\uparrow$   \\ \cline{2-6} 
		\multirow{3}{*}{\rotatebox{90}{\textbf{\footnotesize{Training}}}} 
            &W/out Pretraining      & 0.459  & 0.808   &  0.804 & 0.886   \\
            &W/out NAdam~\cite{dozat2016incorporating}      & 0.096   &  0.965    & 0.966  & 0.992 \\
            &W/out Look-Ahead~\cite{zhang2019lookahead} \phantom{M}   & 0.082   &  0.971    & 0.971 & 0.992 \\ \cline{2-6}
		\multirow{8}{*}{\rotatebox{90}{\textbf{\footnotesize{Data Augmentation}}}}
		&W/out Negative Samples   & 0.319  & 0.905  &0.909  &  0.943\\
		 & W/out Color Aug.   & 0.119  & 0.956 & 0.957 & 0.992  \\
		&W/out Rescales   & 0.105   &  0.960 &  0.968 &   0.990 \\ 
            &W/out Flips   &  0.087  & 0.972  & 0.972  & 0.993 \\ 
		&W/out Distortions   & 0.094  & 0.966 & 0.966 & 0.987 \\ 
		&W/out $AUG_{T \to F}$   & 0.121   & 0.960 & 0.961 & 0.990 \\ 
		  &W/out $AUG_{F \to F'}$   & 0.087 & 0.969 & 0.966 & 0.991 \\ 
		  &W/out UnFold    & 0.082 & 0.970  &  0.971  & 0.992 \\ \cline{2-6} 
            & \textbf{Final Model }  & \textbf{0.064}   & \textbf{0.982}  & \textbf{0.983} & \textbf{0.997} \\ \cline{2-6}
            
	\end{tabular}
        }
        \vspace{-3mm}
	\caption{Ablation study on different configurations of model training configurations and data augmentation policies. }
 \vspace{-7mm}
	\label{tab:ablation}
\end{table}

\subsubsection*{Data Augmentation and Training Configuration}

In Table~\ref{tab:ablation}, we show an ablation study of our data augmentation and optimization setups. From our final model configuration, we remove different components to its training policy to study their impact on generalization. We report different classification metrics measured on our test set, which contains a balanced number of positive and negative examples.

First, we show that the model strongly benefits from pre-training on ImageNet. The model performance can be enhanced by introducing \emph{Nesterov momentum}~\cite{dozat2016incorporating} into the optimizer, and further with Lookahead training~\cite{zhang2019lookahead}.  We observe that these results are consistent across architectures, and that other optimizers, such as RAdam~\cite{liu2019variance} or AdamW~\cite{loshchilov2017decoupled}, performed worse than NAdam or Adam. 

Regarding data augmentation, we first tested a model trained without negative examples, relying on synthetic non-tileable textures from tileable ground truths. However, this yielded limited generalization. Traditional augmentations (color, geometry, noise, blurs, elastic transformations) provided incremental improvements. Our custom policies, which generate negative samples from tileable images and vice versa, further enhanced model performance. By introducing random \emph{unfolding}, we achieve small gains. This comprehensive data augmentation policy mitigates model and dataset limitations and makes TexTile more robust to important factors like color variations, sharpness, or scales.

\begin{table}[t]
	\centering
	\resizebox{0.75\columnwidth}{!}{
	\begin{tabular}{l   @{\extracolsep{\fill}}cccc} Distance & Error $\downarrow$ & Accuracy $\uparrow$ &  $F_1$ $\uparrow$ & AUC $\uparrow$   \\ \cline{1-5} 
            FID~\cite{heusel2017gans} & 0.568 & 0.703  & 0.699 & 0.764  \\
            GS~\cite{khrulkov2018geometry} & 0.694 & 0.507  & 0.533  & 0.510 \\
            MSID~\cite{tsitsulin2019shape}  & 0.797 & 0.582 & 0.544 & 0.503 \\
            \textbf{TexTile}  & \textbf{0.064}   & \textbf{0.982}  & \textbf{0.983} & \textbf{0.997} \\
		
	\end{tabular}
}   \vspace{-3mm}
	\caption{Comparison of our metric with distribution-based distances, on a downstream classification task.}
	\label{tab:comparison_metrics}
 \vspace{-5mm}
\end{table}

\begin{figure*}[tb!]
	\centering
	\includegraphics[width=1.0\textwidth]{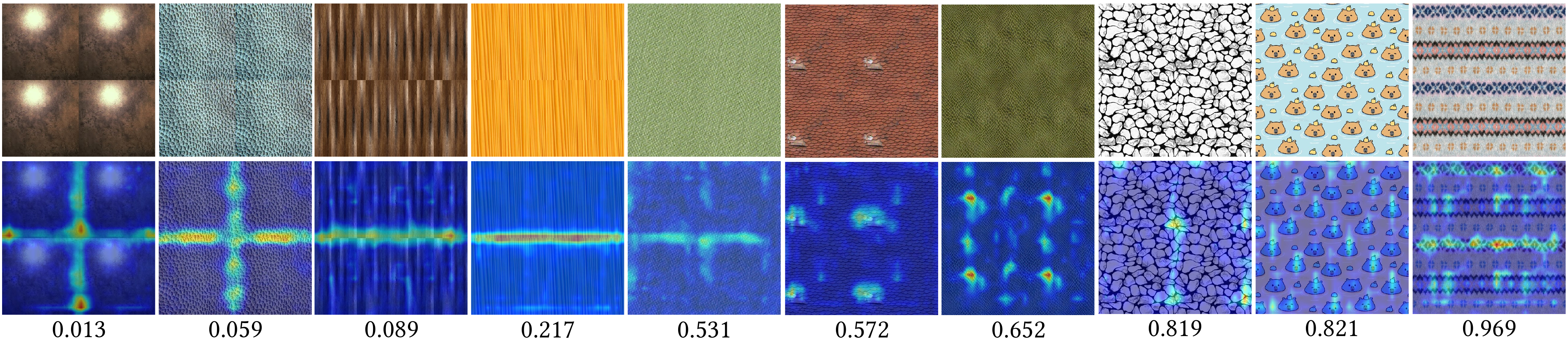}
	\vspace{-7mm}
	\caption{On top, textures samples (tiled $2\times2$) with increasing predicted tileability. Below them, model saliency maps and TexTile values.}
	\label{fig:saliency}
	\vspace{-6mm}
	
\end{figure*}

\subsection{Comparison with Distribution-Based Metrics}
In this experiment, we compare our metric with off-the-shelf distribution-based metrics. %
To achieve this, we compute the latent features of both our tileable and non-tileable training datasets, using 2x2 tilings as in our setup. Then, for randomly selected subsets from our test set, we compare their latent features to those of both training datasets. Finally, we classify the set according to the which of both distributions is closest. We show the results on Table~\ref{tab:comparison_metrics}, where we find that these metrics fail to capture important characteristics that influence the tileability of textures. Directly supervising for tileability is unsurprisingly more effective. Interestingly, the GS~\cite{khrulkov2018geometry} and MSID~\cite{tsitsulin2019shape} metrics perform only marginally better than random, whereas FID~\cite{heusel2017gans} better captures the differences between the distributions.

\subsection{Qualitative Evaluation}\label{sec:qualitative}
We leverage Axion-Based Class-Activation Mappings~\cite{fu2020axiom} to visualize which features are most relevant to our model. In Figure~\ref{fig:saliency}, we show saliency maps and predicted TexTile values for a few representative textures with increasing degrees of tileability. As shown on the first two examples, our model predicts low tileability values for textures without seamless borders. In the next three examples, which are seamless only on one of their axes, the model outputs higher tileability values. On the last five examples, which are all seamlessly tileable, the model is leveraging other features, like uneven shadings, unusual artifacts, or repeating objects.  These results show that our model can exploit patterns other than border discontinuity for its prediction, and that it can integrate distant information for finding repeating elements.   

\section{Results}\label{sec:results}
\subsection{TexTile as a Loss Function}\label{sec:loss}
Because TexTile is a fully-differentiable metric, it can be leveraged as a loss function for synthesizing tileable textures. We explore this on two different types of algorithms.

\begin{figure}[tb!]
	\centering
	\includegraphics[width=1.0\columnwidth]{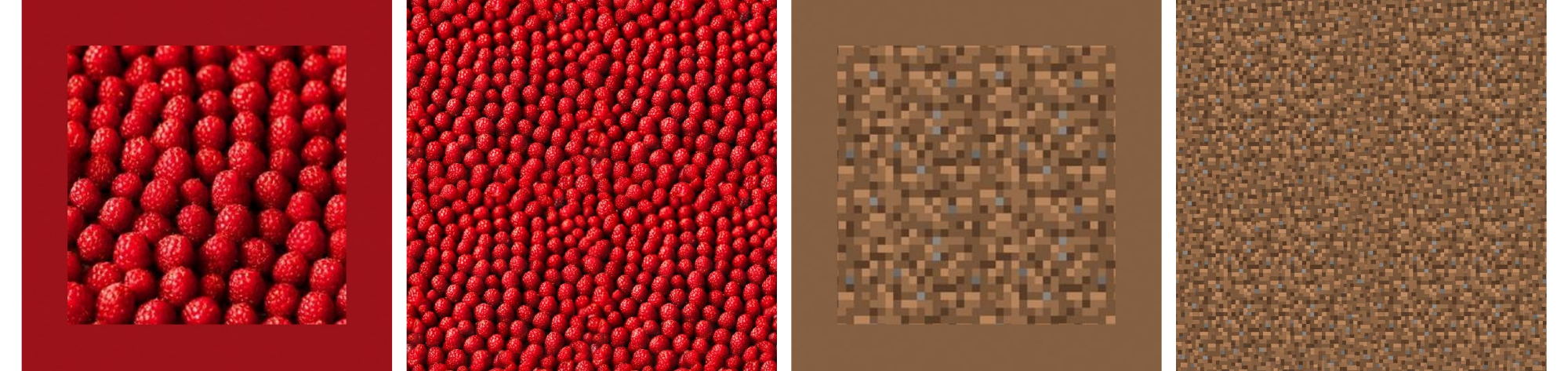}
 \vspace{-5mm}
	\caption{Image outpainting for tileable texture synthesis. On the left, \emph{non-tileable} input images with the area to be outpainted in a solid color; on their sides, outpainted results, obtained by maximizing tileability, shown in a 2x2 tile composition.}
	\label{fig:completion}
 \vspace{-5mm}
\end{figure}

First, we extend an optimization-based \textbf{neural texture synthesis} algorithm~\cite{heitz2021sliced} to generate tileable textures. To do so, we simply optimize a joint loss function $\mathcal{L} = \lambda_{\text{style}} \mathcal{L}_{\text{style}} + \lambda_{\text{TexTile}} \mathcal{L}_{\text{TexTile}}$, where $\lambda_{\text{style}}$ and $\lambda_{\text{TexTile}}$ control the weight of the each component of the loss and are selected empirically to
$\lambda_{\text{style}} = \lambda_{\text{TexTile}} = 1$. We observed little sensitivity to these weightings as long as they are on the same order of magnitude. %
We show end-to-end synthesis results in Figure~\ref{tab:comparisons}. We can also generate tileable textures in this fashion by optimizing only the texture borders using outpainting, as we illustrate in Figure~\ref{fig:completion}. 

Relatedly, we can also extend \textbf{single image diffusion models} so they can generate tileable textures. While the goal of these methods is more general image or video synthesis, we leverage them as powerful texture synthesis algorithms. To do so, we use SinFusion models~\cite{nikankin2022sinfusion} without any modifications in the training process. During inference, after each diffusion step, we perform a single optimization step to the noisy image on the direction that maximizes TexTile.  We show qualitative results in Figure~\ref{tab:comparisons}. 

 With these simple modifications, we can transform these methods --and potentially any texture generative model-- into tileable texture synthesis algorithms, without significant loss in the perceptual quality of the generated textures. %
 Importantly, this can be achieved without re-training the generative models.  Further implementation details and results are included in the supplementary material.

\subsection{Benchmarking Texture Synthesis Algorithms}\label{sec:benchmark}
In Table~\ref{tab:quantitative_methods}, we show a quantitative comparison between different texture synthesis algorithms and generative models, on the 14-texture dataset used in~\cite{rodriguezseamlessgan}, using reference and no-reference metrics.  For~\cite{heitz2021sliced,nikankin2022sinfusion}, we show results of their baseline methods and our modifications that generate tileable textures. The qualitative results on the complete dataset is present in the supplementary material.  %

Besides, we observe that powerful generative models like~\cite{rodriguezseamlessgan,nikankin2022sinfusion} do not consistently beat non-parametric alternatives like~\cite{deliot2019procedural,li2020inverse} across either reference or no-reference metrics. In terms of tileability, introducing Textile as a loss function to~\cite{heitz2021sliced,nikankin2022sinfusion} not only significantly improves the tileability of their outputs but it does it without reducing their perceptual quality. %
Unsurprisingly, methods that perform border blending~\cite{deliot2019procedural,rodriguez2019automatic} achieve lower tileability values across different metrics, than methods that synthesize the textures in a more holistic way, like~\cite{moritz2017texture,rodriguezseamlessgan,niklasson2021self-organising,bergmann2017learning}. These results confirm that there are more constituent factors in texture tileability than simply seamless borders. 

\begin{figure}[tb!]
	\centering
	\includegraphics[width=1.0\columnwidth]{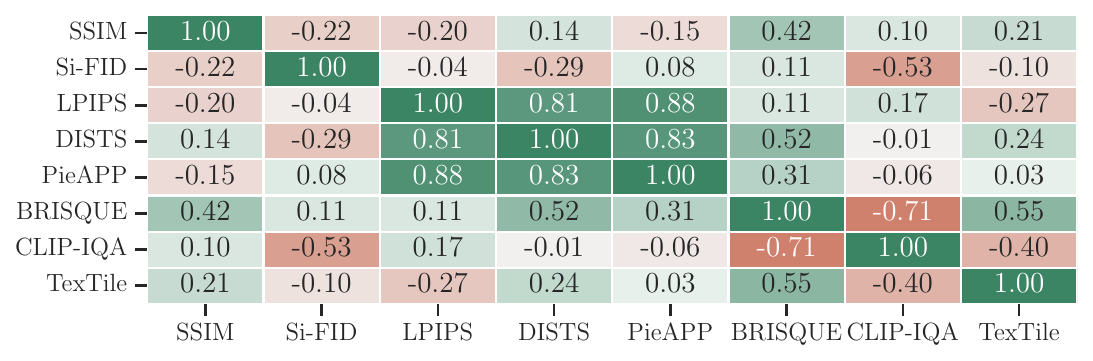}
 \vspace{-6mm}
	\caption{Pearson correlation matrix between different metrics.}
		\vspace{-3mm}
	\label{fig:correlations}
\end{figure}

A correlation matrix between these metrics is shown in Figure~\ref{fig:correlations}. %
Learned reference metrics (LPIPS, DISTS and PieAPP) correlate strongly between each other but poorly with other measures. Si-FiD and SSIM are not closely related with any other metric, while TexTile is slightly correlated with BRISQUE and CLIP-IQA. As we also illustrate in Figure~\ref{fig:teaser}, there is no consensus across metrics on which algorithm is outperforming the others, showing that perceptual evaluation for texture synthesis remains a challenge.

\begin{table*}[]
	\centering
        \resizebox{\textwidth}{!}{
	\begin{tabular}{@{}lccccc|ccc@{}}
		\cmidrule(l){2-9} 
		\multicolumn{1}{c}{\multirow{2}{*}{}} & \multicolumn{5}{c}{\textbf{Reference-Based Metrics}}  & \multicolumn{3}{c}{\textbf{No Reference}} \\ \cmidrule(l){2-9} 
		\small
		
		&SSIM~\cite{wang2004image} $\uparrow$&Si-FID~\cite{rottshaham2019singan} $\downarrow$ &LPIPS~\cite{zhang2018unreasonable} $\downarrow$  & DISTS~\cite{ding2020image} $\downarrow$  & PieAPP~\cite{prashnani2018pieapp} $\downarrow$ & BRISQUE~\cite{mittal2012no} $\downarrow$  & CLIP-IQA~\cite{wang2023exploring} $\uparrow$ & \textbf{TexTile}  $\uparrow$ \\  \cmidrule(l){2-9} 
		Deloit \etal~\cite{deliot2019procedural} & 0.138 & 1.707 & 0.595 & 0.372 & 2.289 & 49.43 & 0.407   & 0.639  \\
		Li \etal ~\cite{li2020inverse} &0.149 & 0.981 & \textbf{0.559} & 0.341 & 1.672 & 45.84 & 0.437   & 0.707 \\
		
		Rodriguez-Pardo \etal~\cite{rodriguez2019automatic}&0.171 & 0.969 & 0.594 & 0.361 & 1.890 & \textbf{43.14} & \textbf{0.661}   & 0.403 \\ 
		Bergmann \etal~\cite{bergmann2017learning}  &0.148 & 0.981 & 0.579 & 0.359 & 1.789 & \textbf{44.13} & \textbf{0.638} & 0.675\\ 
		Niklasson \etal~\cite{niklasson2021self-organising}  &0.146 & \textbf{0.819} & 0.623 & 0.446 & 2.292 & 53.99 & 0.435   & 0.731 \\ 
		Rodriguez-Pardo \etal~\cite{rodriguezseamlessgan}    &0.166 & \textbf{0.694} & \textbf{0.540} & 0.336 & \textbf{1.542} & 51.10 & 0.452  & 0.729 \\ \cmidrule(l){2-9} 
        Heitz \etal~\cite{heitz2021sliced} w/out TexTile &0.140 & 1.741 & 0.575 & \textbf{0.329} & 1.687 & 49.05 & 0.376  & 0.431 \\
        Heitz \etal~\cite{heitz2021sliced} with TexTile &0.152 & 1.764 & 0.555 & \textbf{0.327} & \textbf{1.653} & 48.49 & 0.392  & \textbf{0.781} \\ \cmidrule(l){2-9} 
        Nikankin \etal~\cite{nikankin2022sinfusion} w/out TexTile &\textbf{0.172} & 1.314 & 0.591 & 0.386 & 1.963 & 55.96   & 0.408   & 0.388 \\
        Nikankin \etal~\cite{nikankin2022sinfusion} with TexTile & 
        \textbf{0.189} & 1.415 & 0.569 & 0.387 & 1.926 & 56.99 & 0.396   & \textbf{0.798} \\ \bottomrule
	\end{tabular}
        }
	\caption{Quantitative evaluation between different tileable texture synthesis algorithms across a variety of metrics, including reference and no-reference measures, and TexTile.  Best two results for each columns are marked in \textbf{bold}. }
	\label{tab:quantitative_methods}
 \vspace{-3mm}

\end{table*}

\makeatletter

\newcommand{\addpic}{\includegraphics[width=0.10\textwidth]{example-image}}
\newcolumntype{C}{>{\centering\arraybackslash}m{.105\textwidth}}
\begin{table*}[t]
	\renewcommand{\@captype}{figure}
	\footnotesize{
	\centering
	\begin{tabular}{C*7{C}@{}}
		\cmidrule[0.1em](l){2-8}
		\multicolumn{1}{c}{\multirow{2}{*}{}} & \multicolumn{3}{c}{\textbf{Previous Methods}}  & \multicolumn{2}{c}{\textbf{Neural Texture Synthesis}~\cite{heitz2021sliced}} & \multicolumn{2}{c}{\textbf{Diffusion Models}~\cite{nikankin2022sinfusion}}  \\ \cmidrule[0.07em](lr){2-4}  \cmidrule[0.07em](lr){5-6}  \cmidrule[0.07em](l){7-8}  
		\textbf{Input} & Blending~\cite{deliot2019procedural} & SeamlessGAN~\cite{rodriguezseamlessgan} & Self-Org~\cite{niklasson2021self-organising} &  W/out TexTile &  With TexTile  & W/out TexTile  & With TexTile    \\ 
		\cmidrule[0.12em](l){1-8}  
		\includegraphics[width=0.105\textwidth]{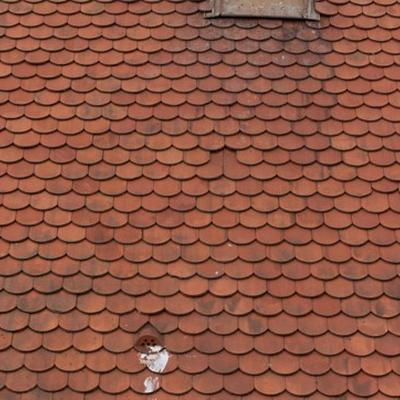}& \includegraphics[width=0.105\textwidth]{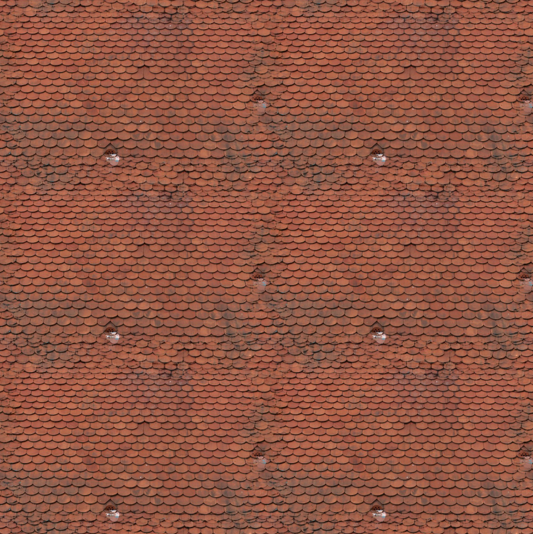} &
		\includegraphics[width=0.105\textwidth]{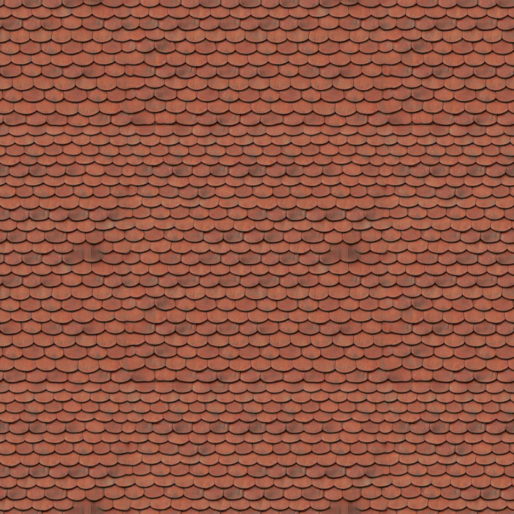} & \includegraphics[width=0.105\textwidth]{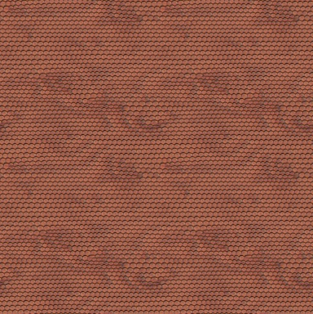} & \includegraphics[width=0.105\textwidth]{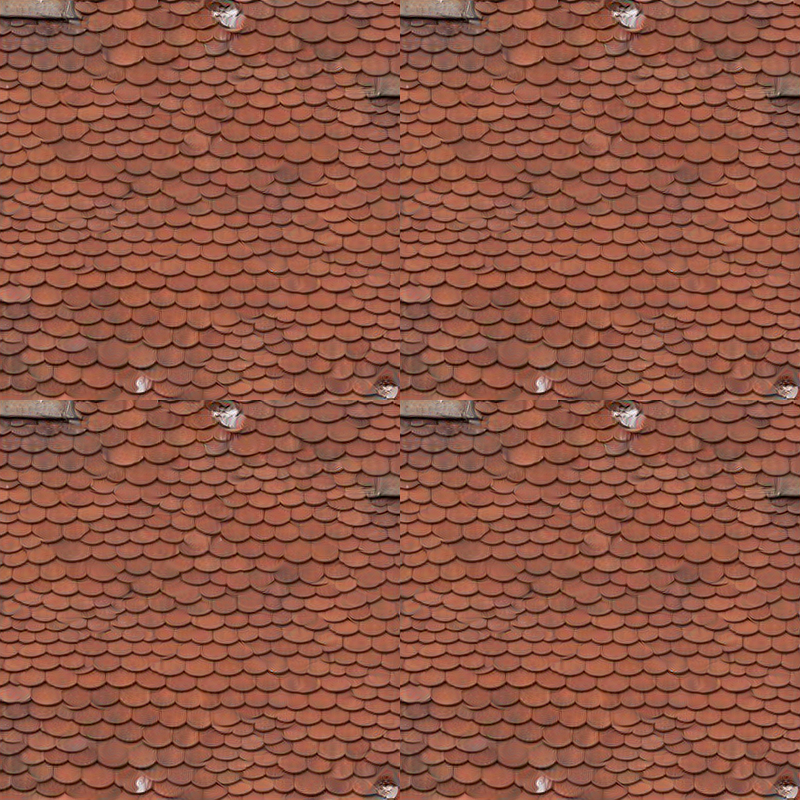} & \includegraphics[width=0.105\textwidth]{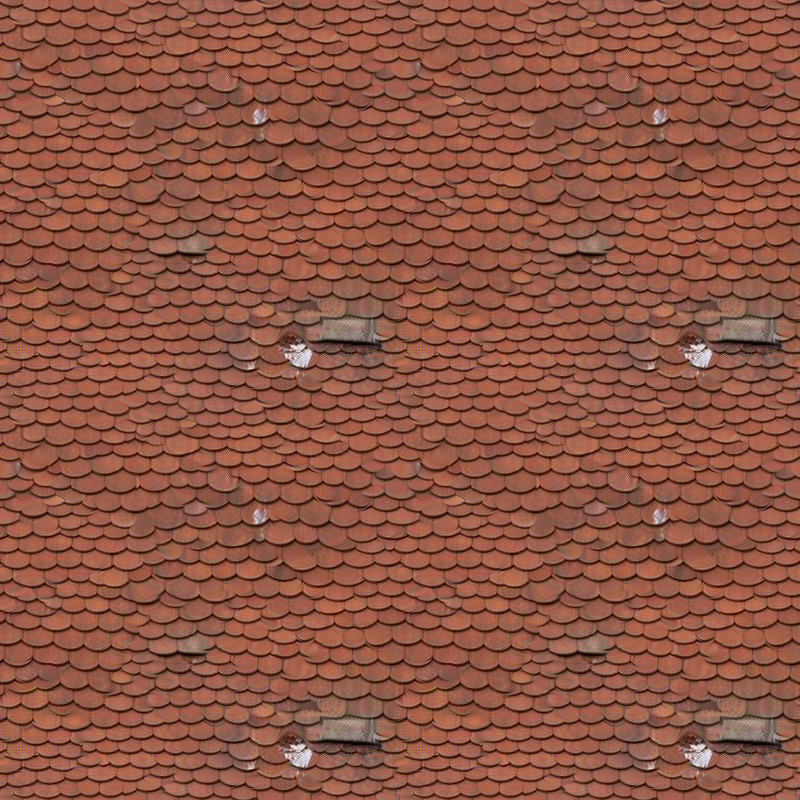}  & \includegraphics[width=0.105\textwidth]{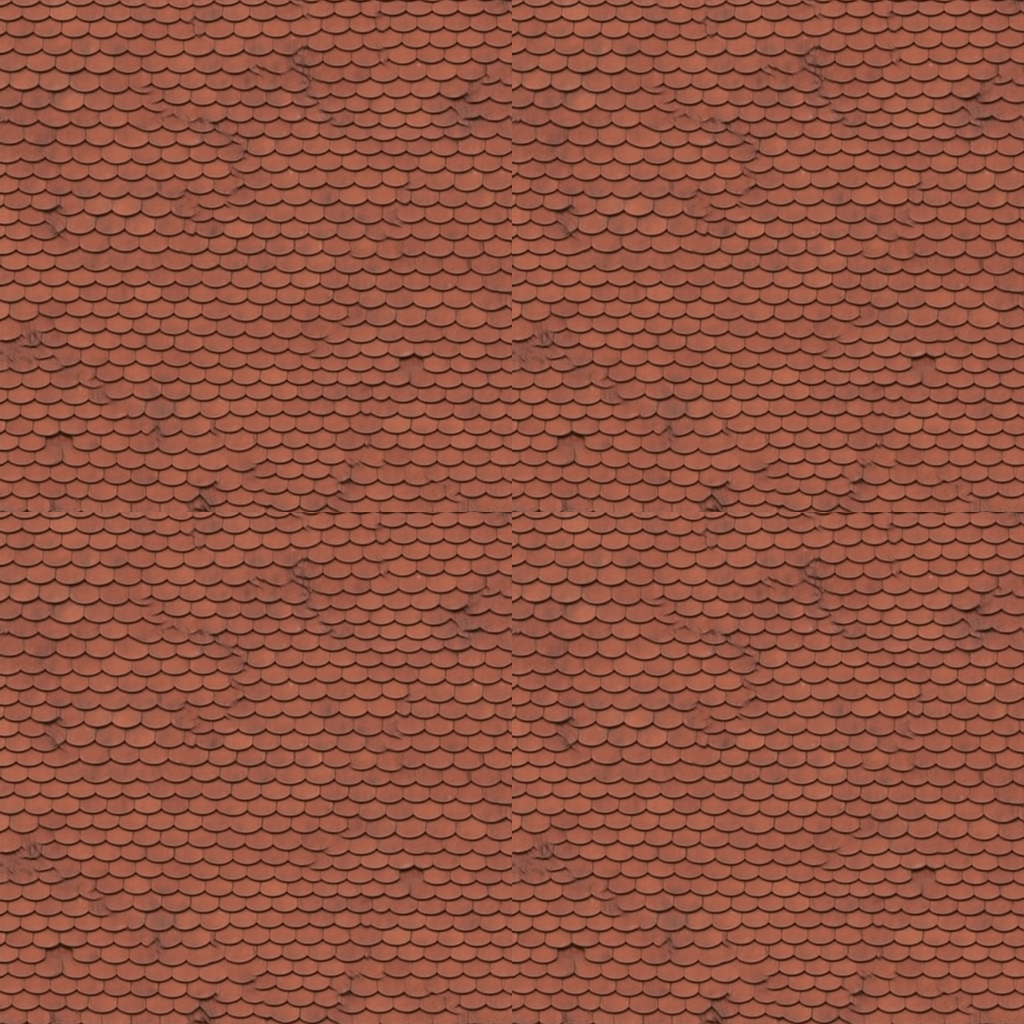} &
		\includegraphics[width=0.105\textwidth]{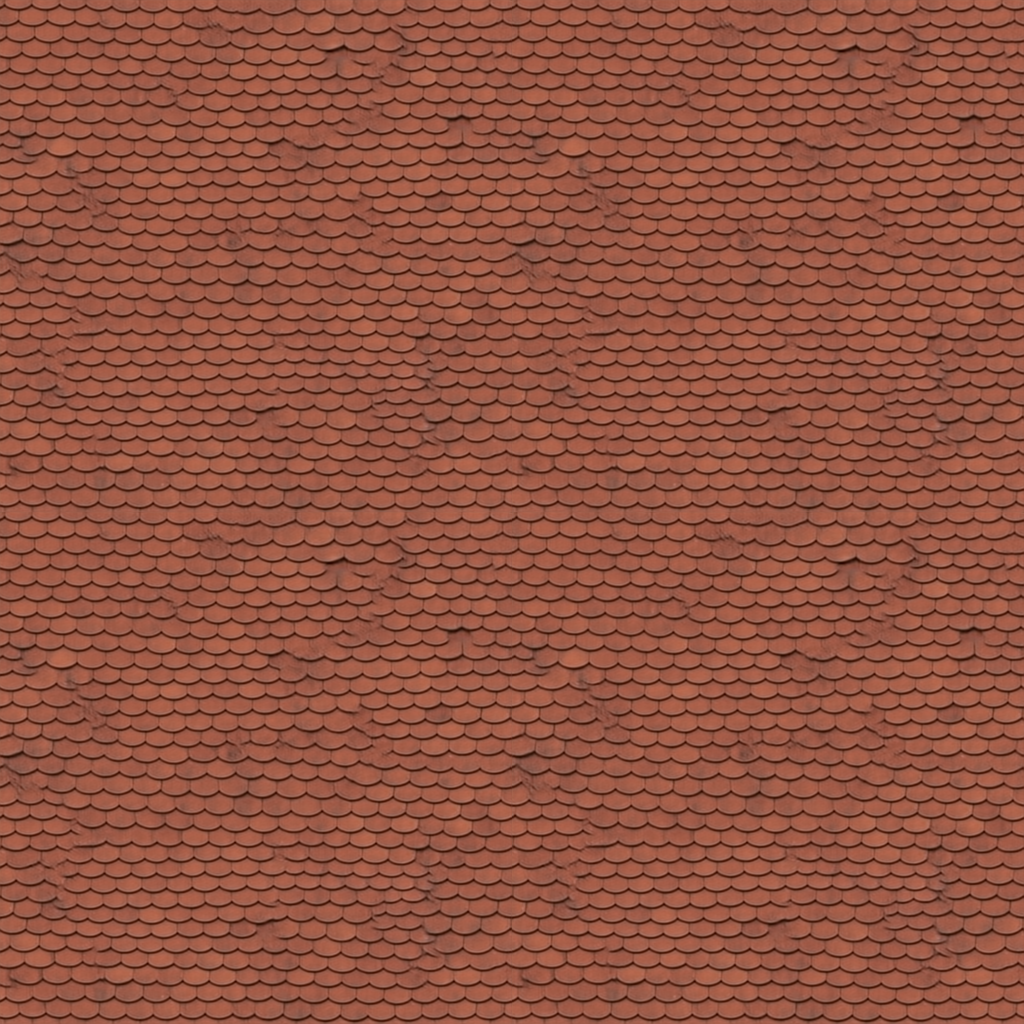}   \\ 
		\includegraphics[width=0.105\textwidth]{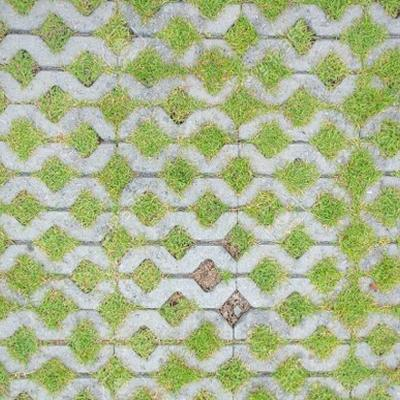}&
		\includegraphics[width=0.105\textwidth]{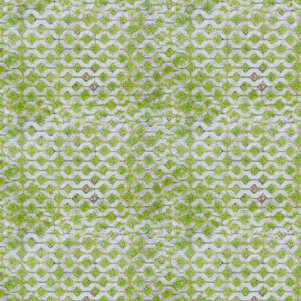}& \includegraphics[width=0.105\textwidth]{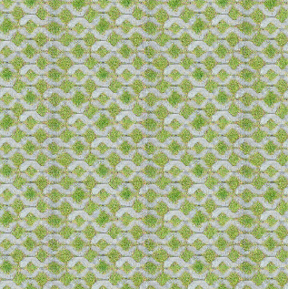} & \includegraphics[width=0.105\textwidth]{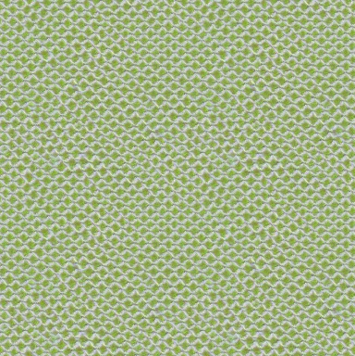} & \includegraphics[width=0.105\textwidth]{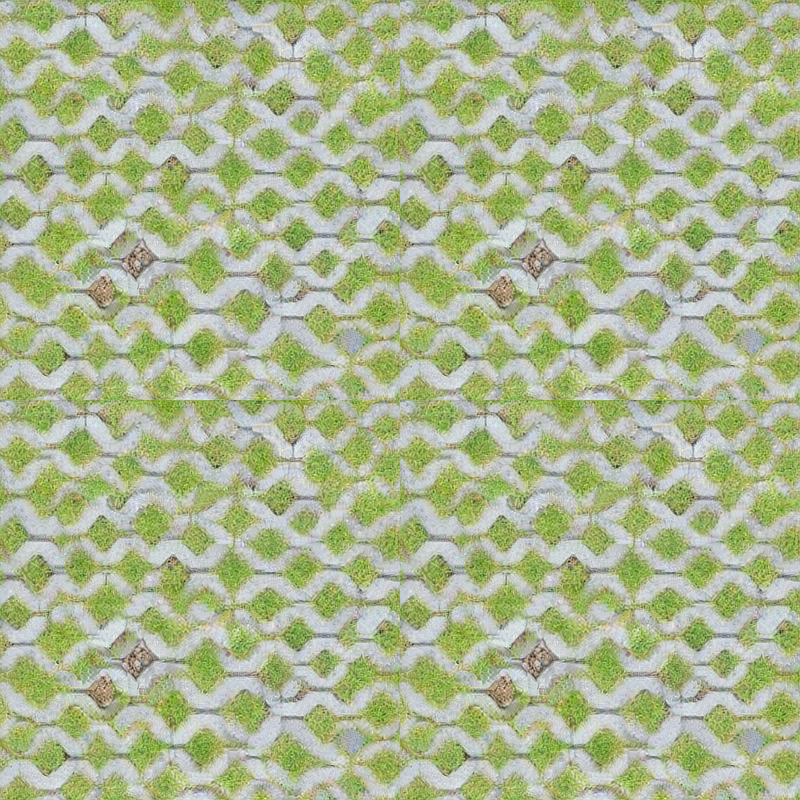} & \includegraphics[width=0.105\textwidth]{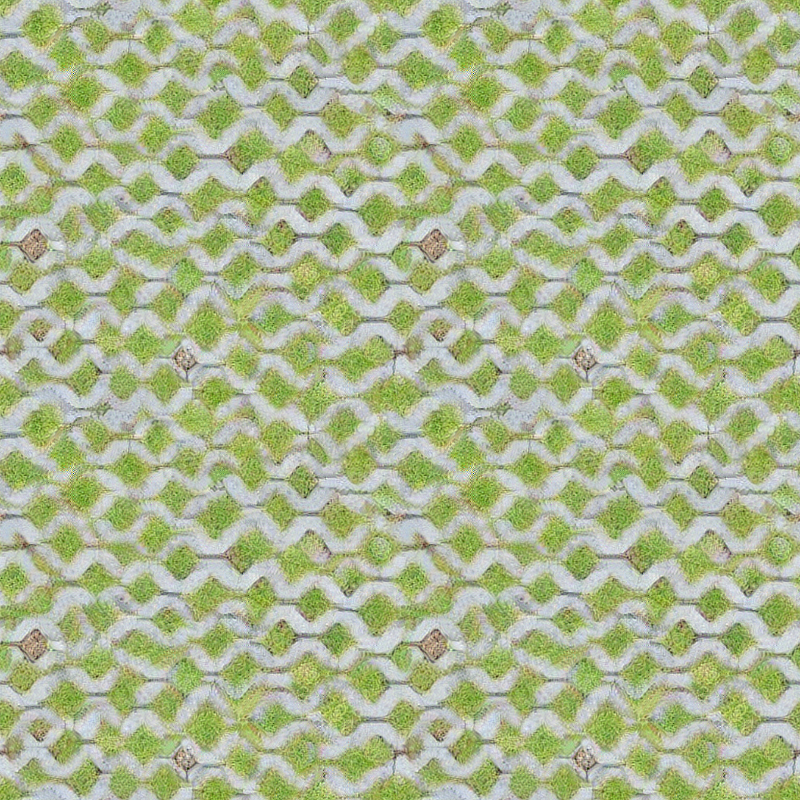}  & \includegraphics[width=0.105\textwidth]{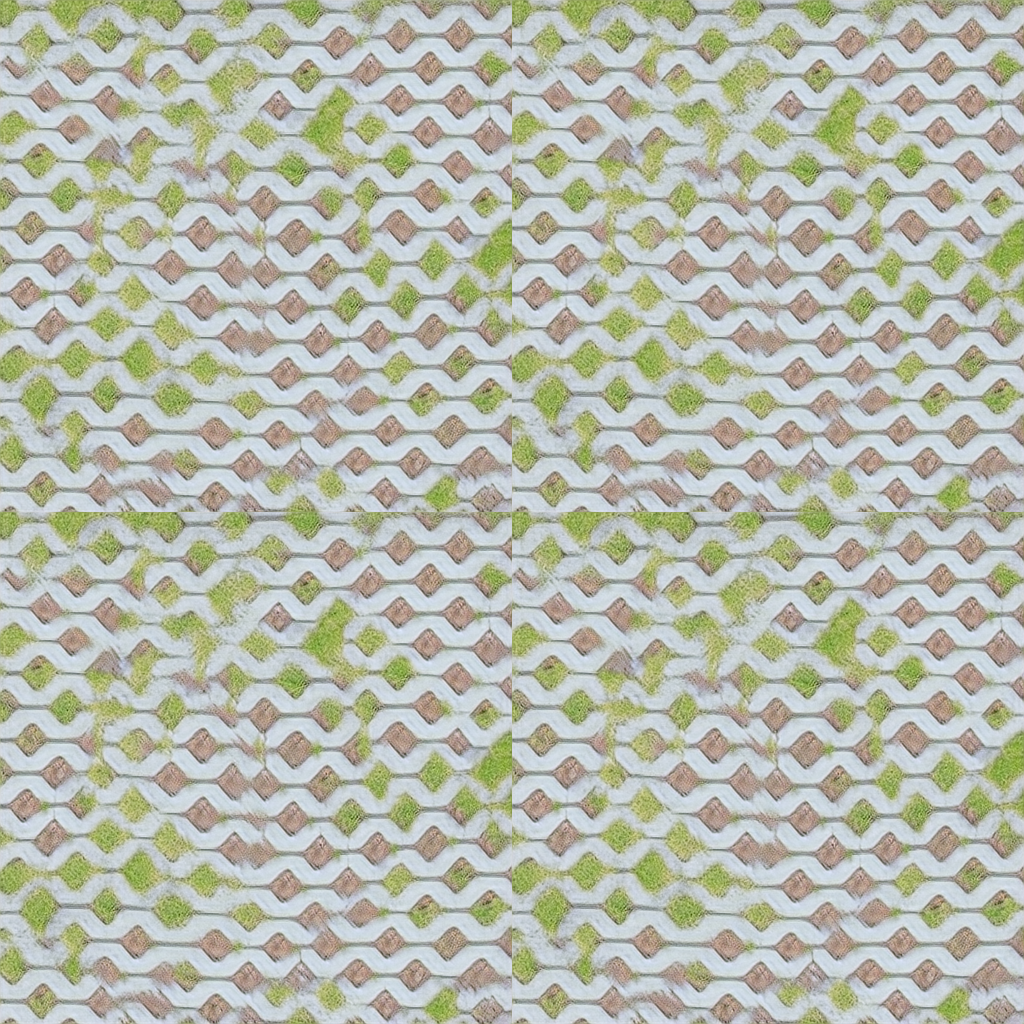} &
		\includegraphics[width=0.105\textwidth]{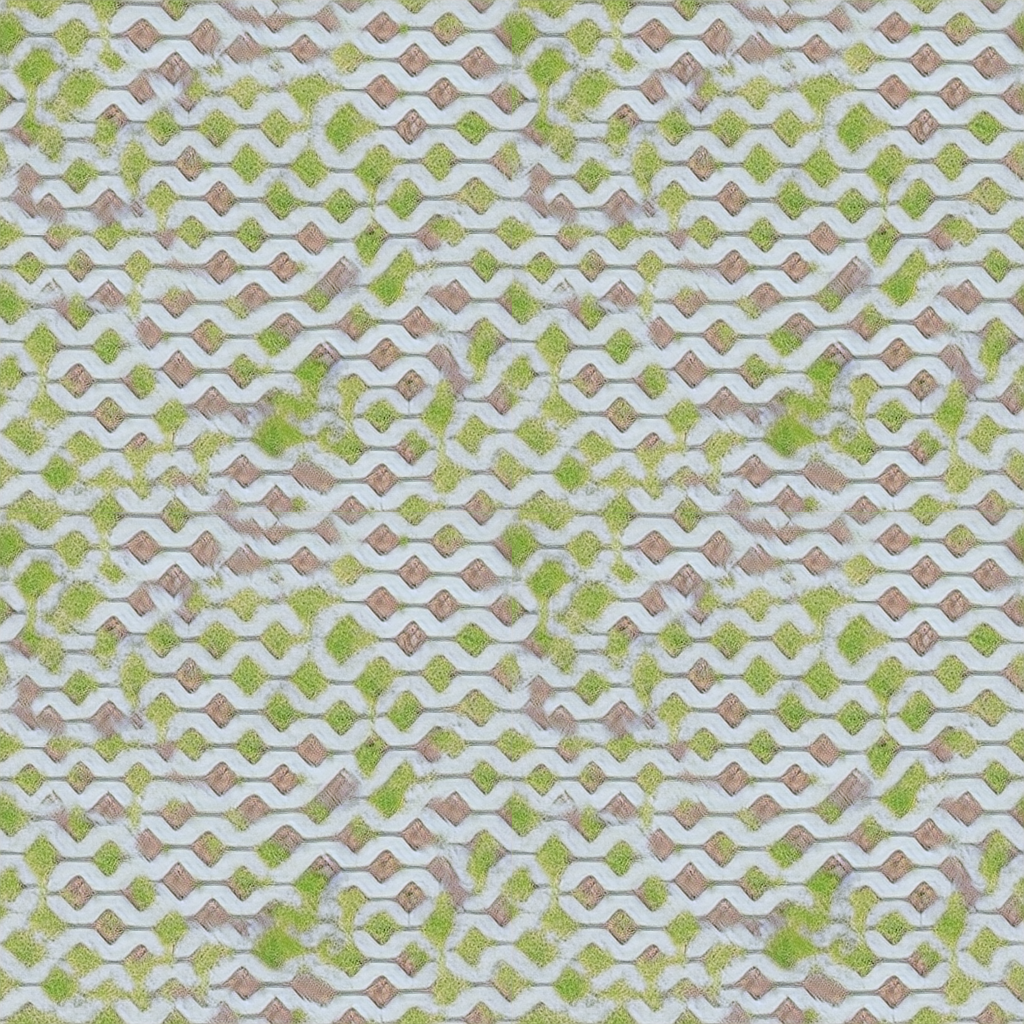}   \\ 
		\includegraphics[width=0.105\textwidth]{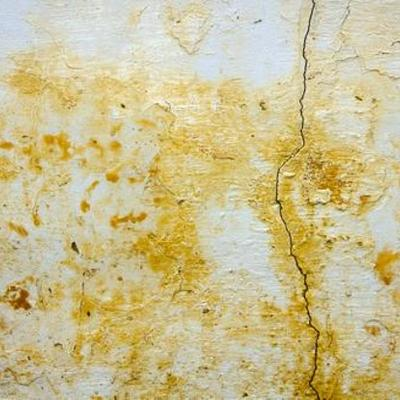}&
		\includegraphics[width=0.105\textwidth]{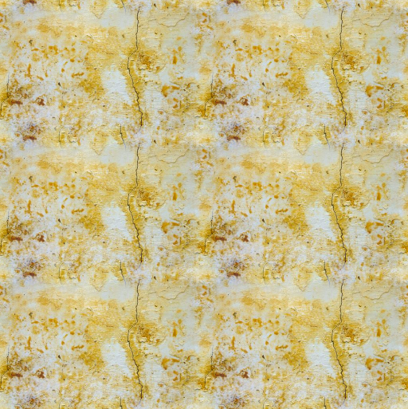}& \includegraphics[width=0.105\textwidth]{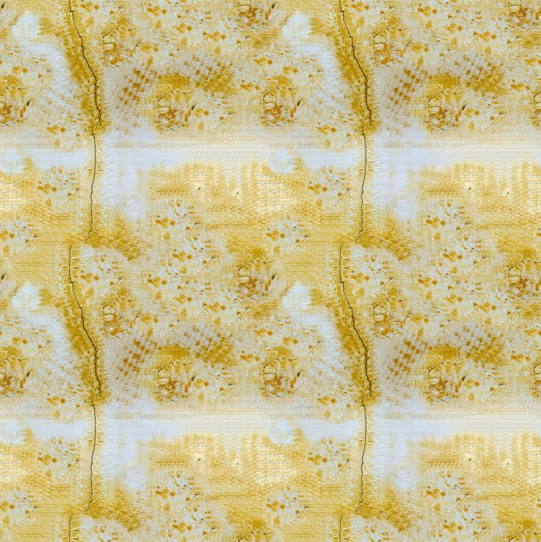} & \includegraphics[width=0.105\textwidth]{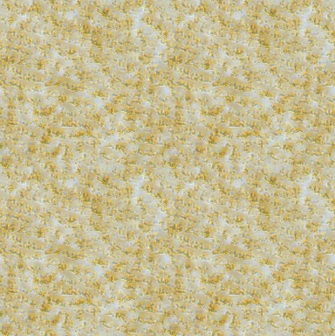} & \includegraphics[width=0.105\textwidth]{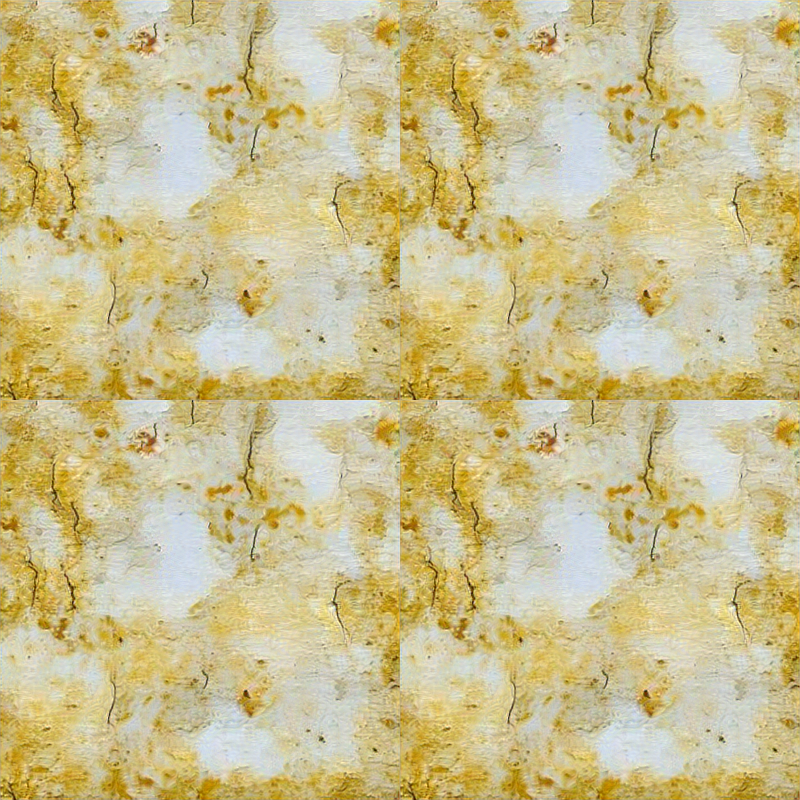} & \includegraphics[width=0.105\textwidth]{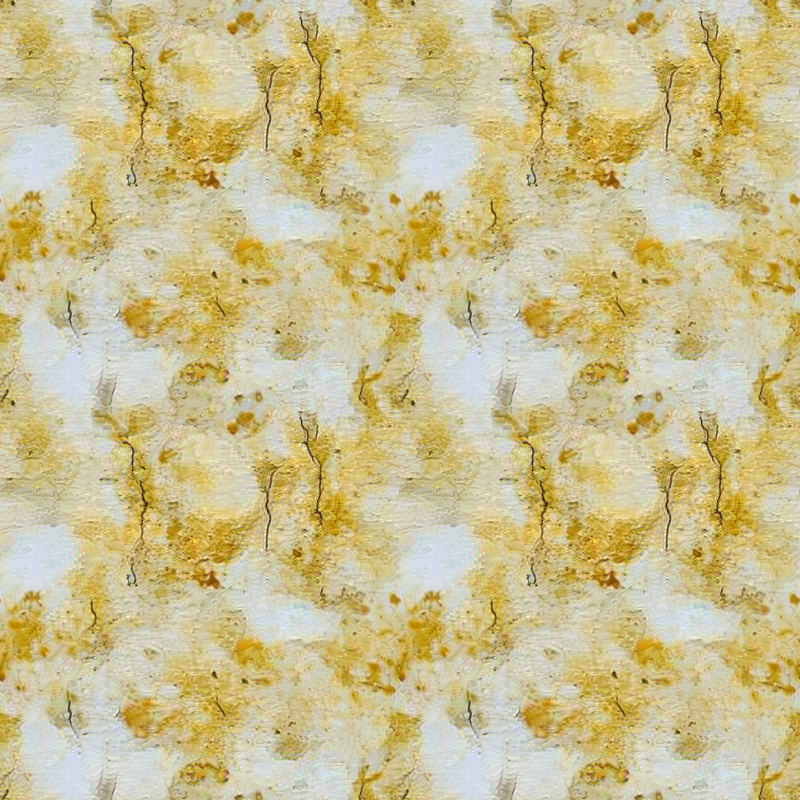}  & \includegraphics[width=0.105\textwidth]{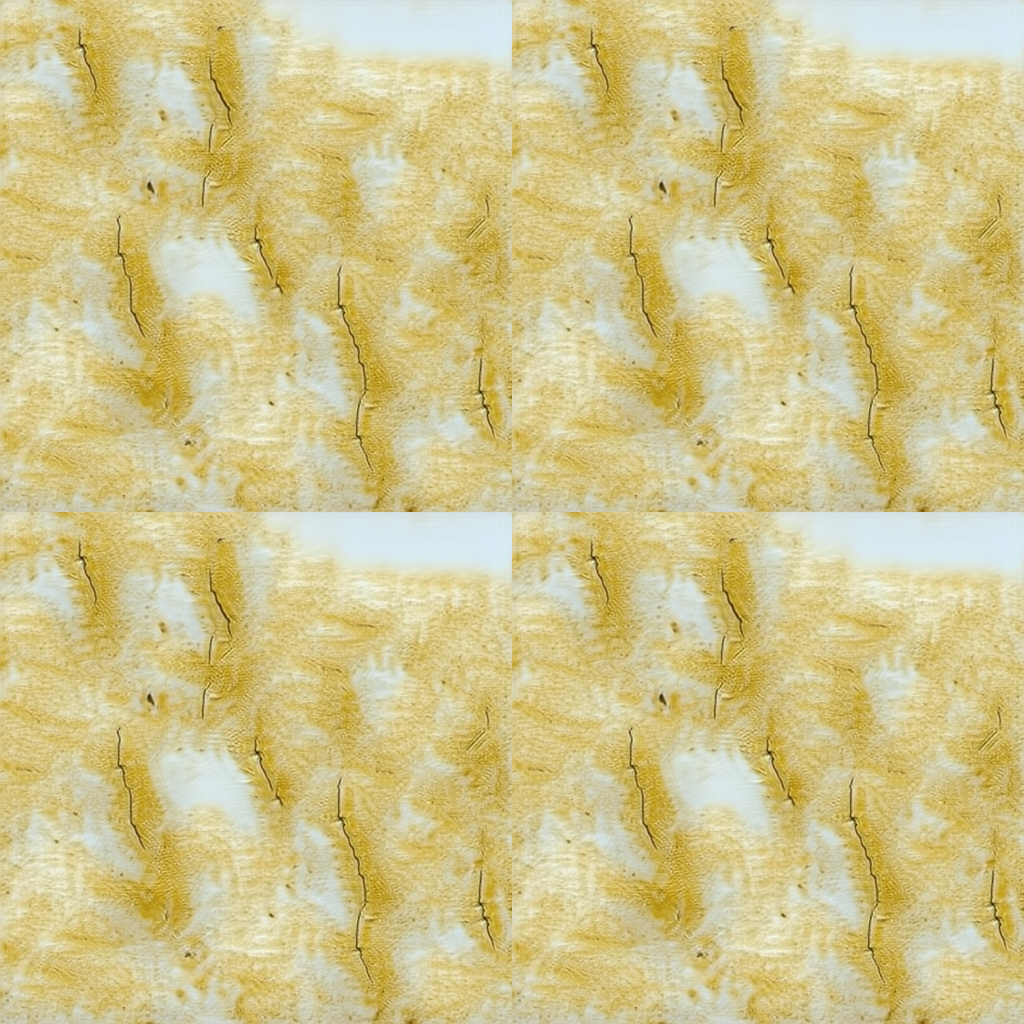} &
		\includegraphics[width=0.105\textwidth]{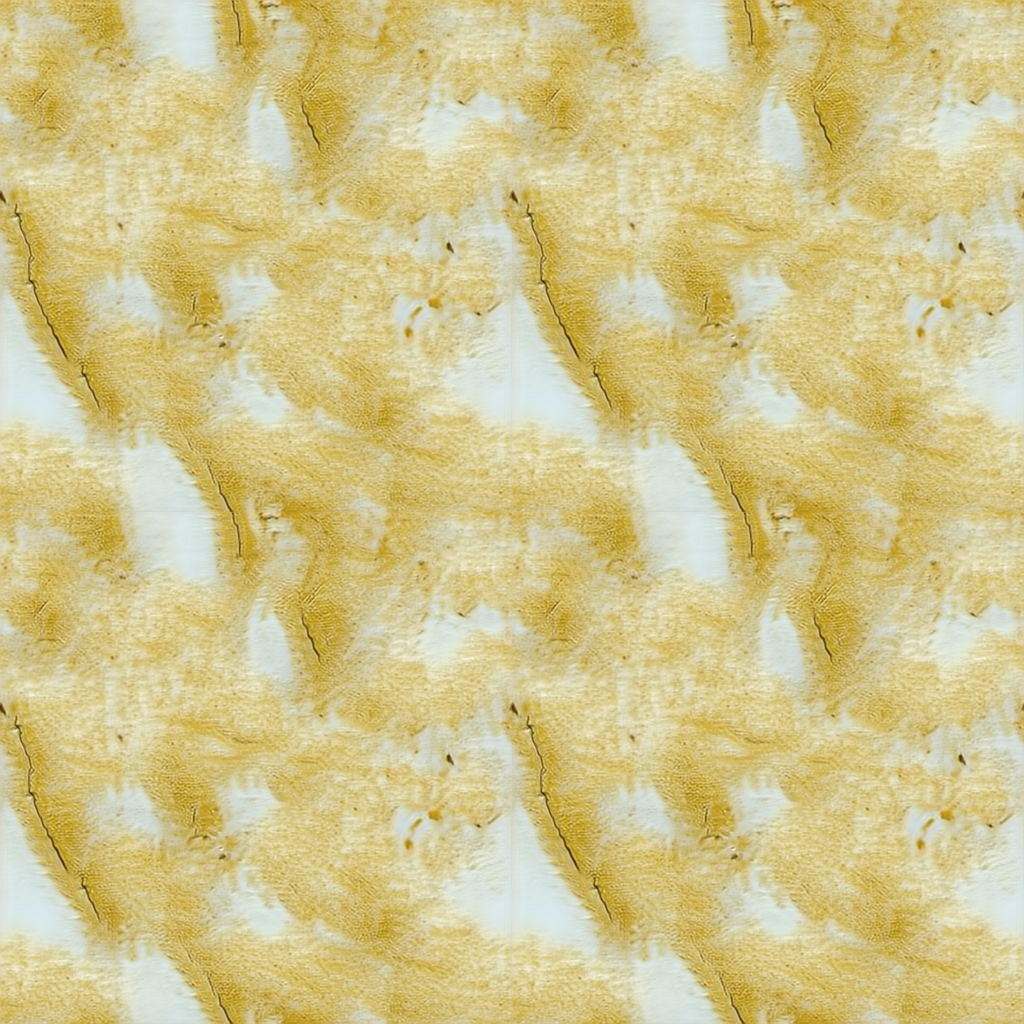}   \\ 
		\includegraphics[width=0.105\textwidth]{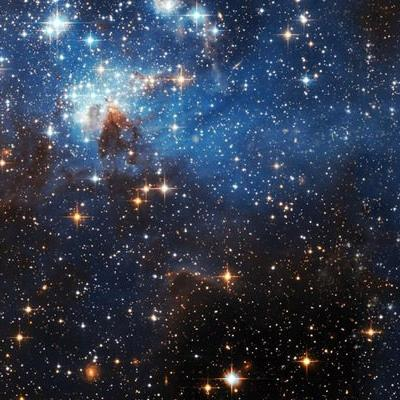}&
		\includegraphics[width=0.105\textwidth]{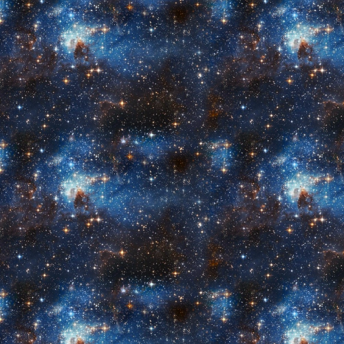}& \includegraphics[width=0.105\textwidth]{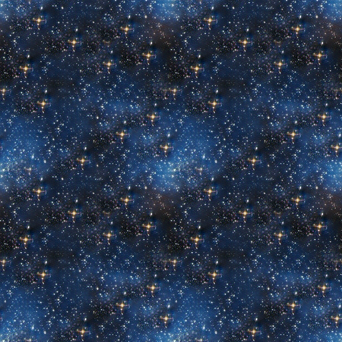} & \includegraphics[width=0.105\textwidth]{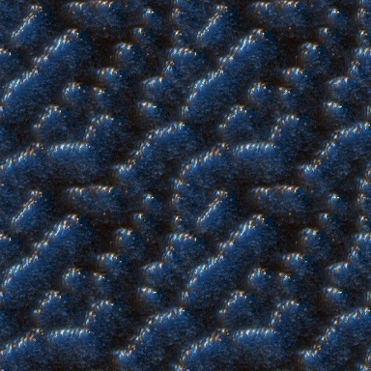} & \includegraphics[width=0.105\textwidth]{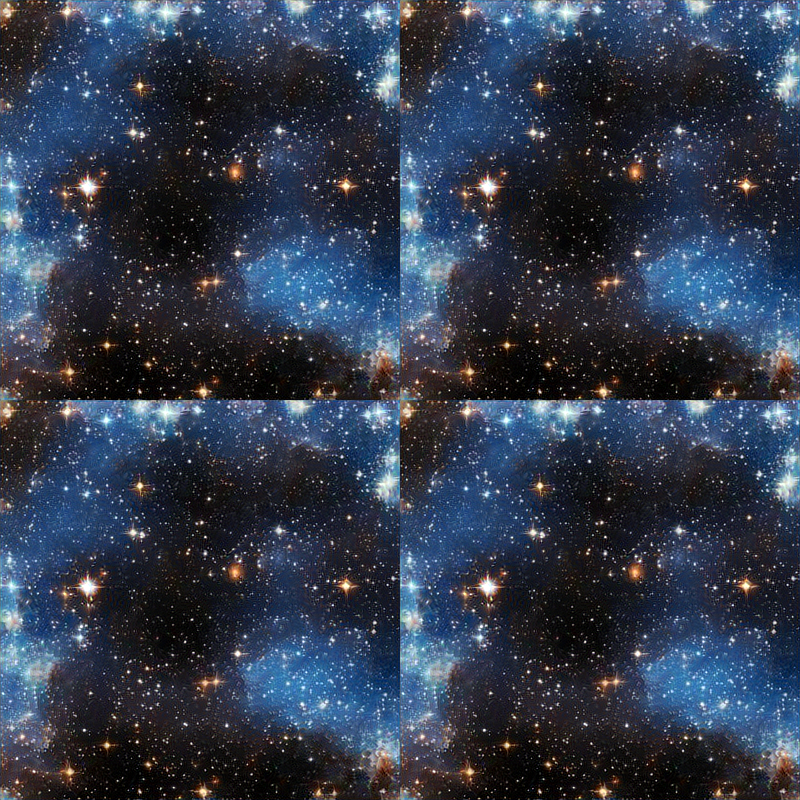} & \includegraphics[width=0.105\textwidth]{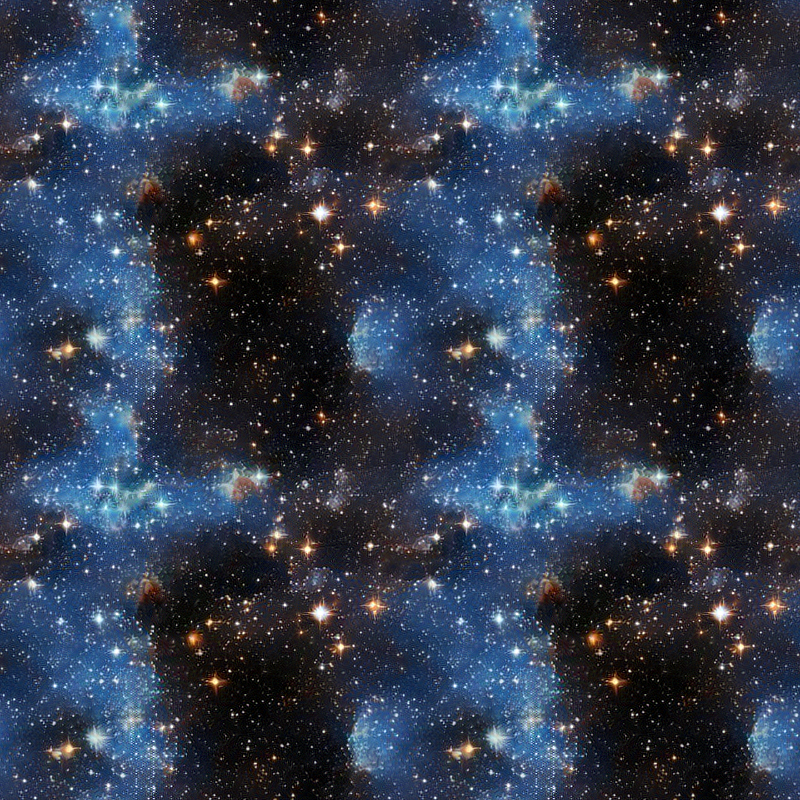}  & \includegraphics[width=0.105\textwidth]{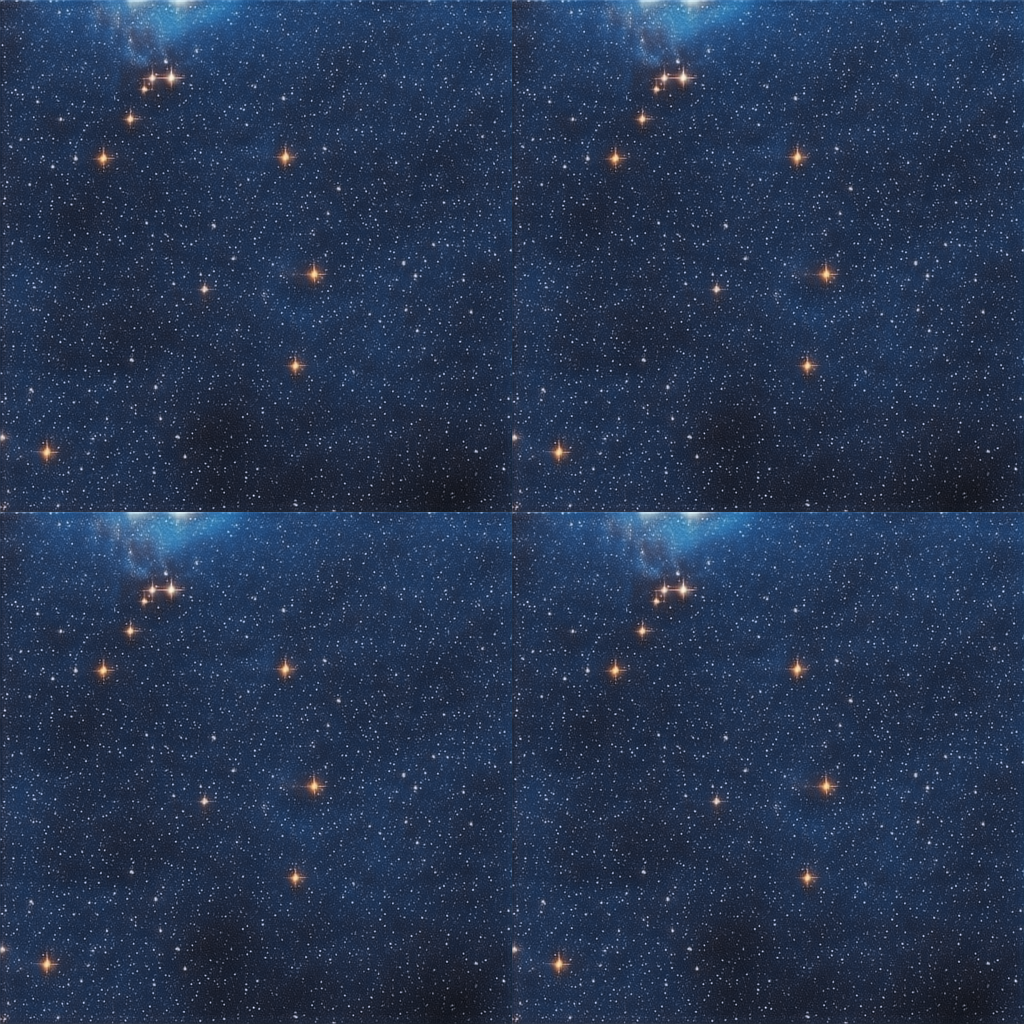} &
		\includegraphics[width=0.105\textwidth]{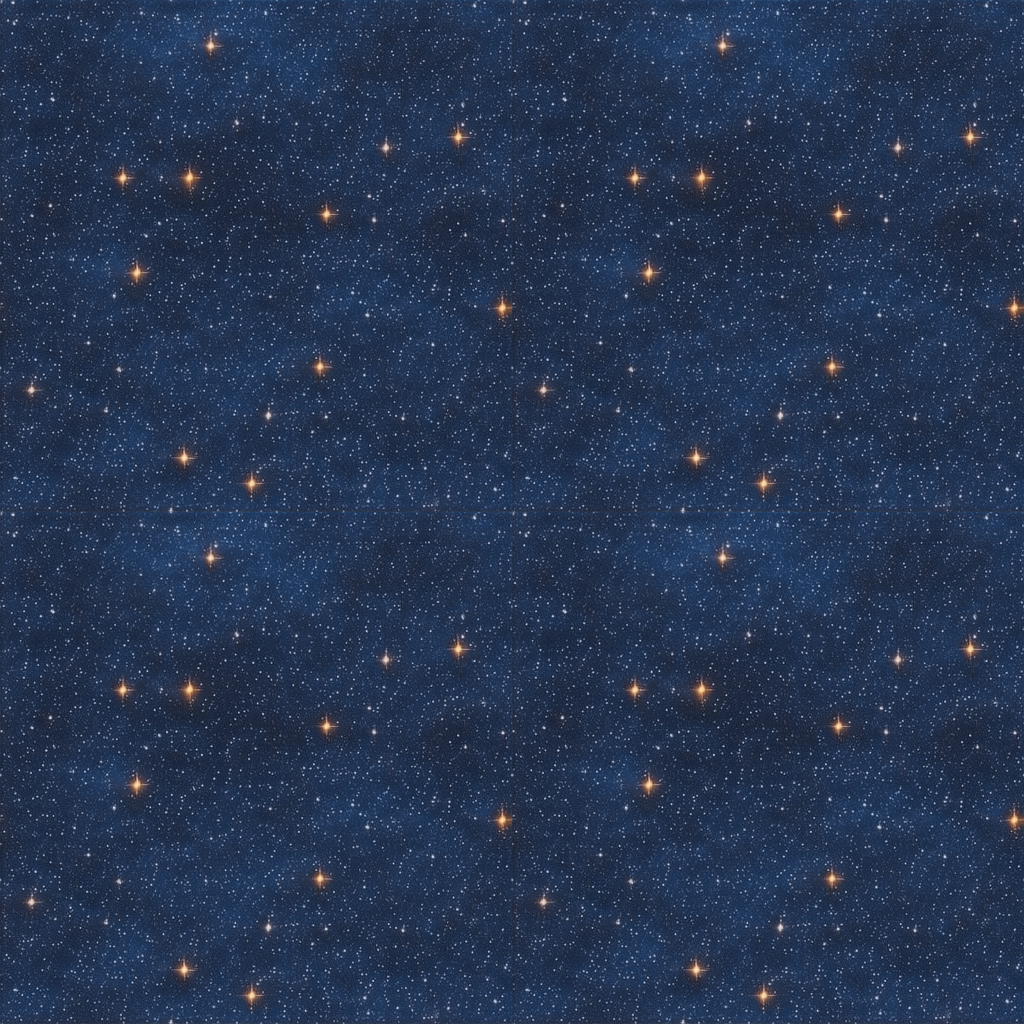}   \\ 
		
		\cmidrule[0.12em](l){1-8}  
	\end{tabular}
	}
 \vspace{-4mm}
	\caption{Comparisons of different texture synthesis algorithms. On the leftmost column, we show the input texture, on the right, 2x2 tilings of the outputs of different methods. For~\cite{heitz2021sliced} and~\cite{nikankin2022sinfusion}, we show the original versions our the modifications for tileable texture synthesis. }
	\label{tab:comparisons}

\end{table*} 
\makeatother

\subsection{Alignment and Repeating Pattern Detection}\label{sec:other_applications}

Besides benchmarking and enabling tileable texture synthesis, our metric enables additional applications. 
\begin{figure}%
	\centering
	\subfloat{{\includegraphics[width=0.51\columnwidth]{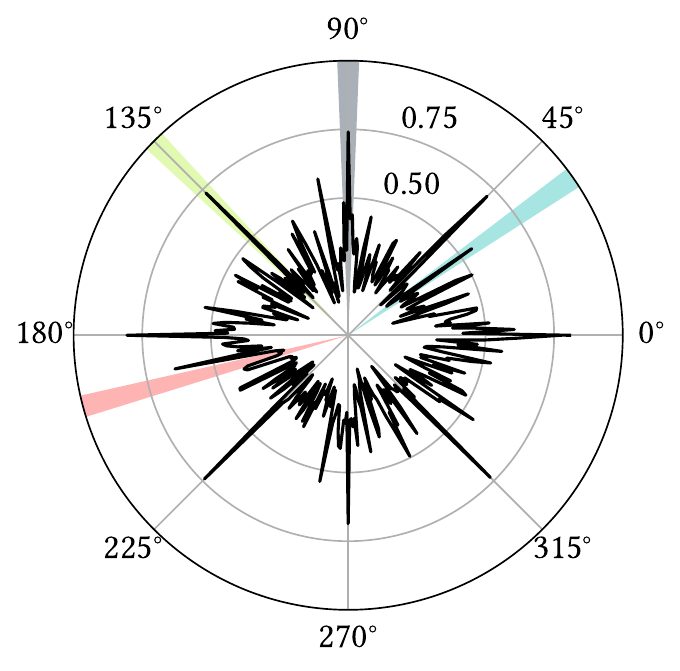} }}%
	\hfil
	\subfloat{{\includegraphics[width=0.46\columnwidth]{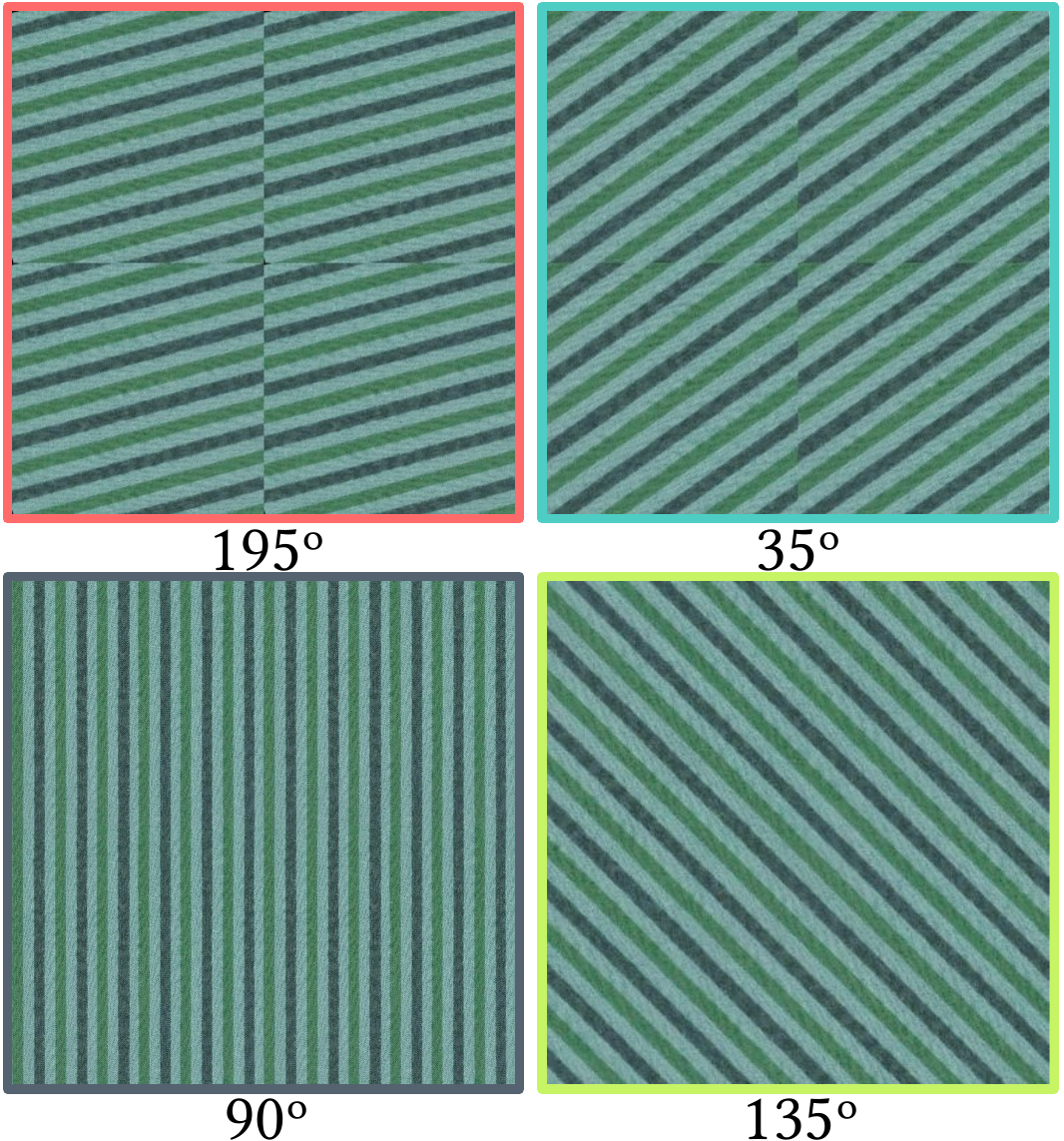} }}%
	\caption{On the left, TexTile under different rotation angles. On the right, samples of rotated images on different local peaks. Scores below 0.5, as the red peak, depict non-tileable textures with highly-noticeable artifacts. A local maxima for a non-tileable texture is found at $\pm35^\circ$, in blue, while highly tileable textures are found at the gray and green insets.}%
	\label{fig:alignment}%
	\vspace{-5mm}
\end{figure}

Previous work on texture analysis used the Radon Transform~\cite{jafari2005radon,rodriguez2019automatic} for automatically aligning images with the $xy$-axes. Similarly, we can find the \textbf{optimal rotation angle} $\theta$ for an image $\text{I}$ by $\argmax_\theta \text{TexTile}(\textrm{Rotate}(\text{I}, \theta))$, maximizing the tileability of the input image. In Figure~\ref{fig:alignment}, we show the scores of our metric, for the same image, to which we apply different rotation angles. Our metric provides high scores for rotation angles which provide seamless borders, and lower values for misalignments. Interestingly, there is a small peak at $\pm35^\circ$, in which the lines in the image connect with each other, but their colors do not match in the borders. These results indicate that TexTile not only measures seamlessness, but also color continuity, and axis alignment. %

\begin{figure}%
	\centering
	\subfloat{{\includegraphics[width=0.275\columnwidth]{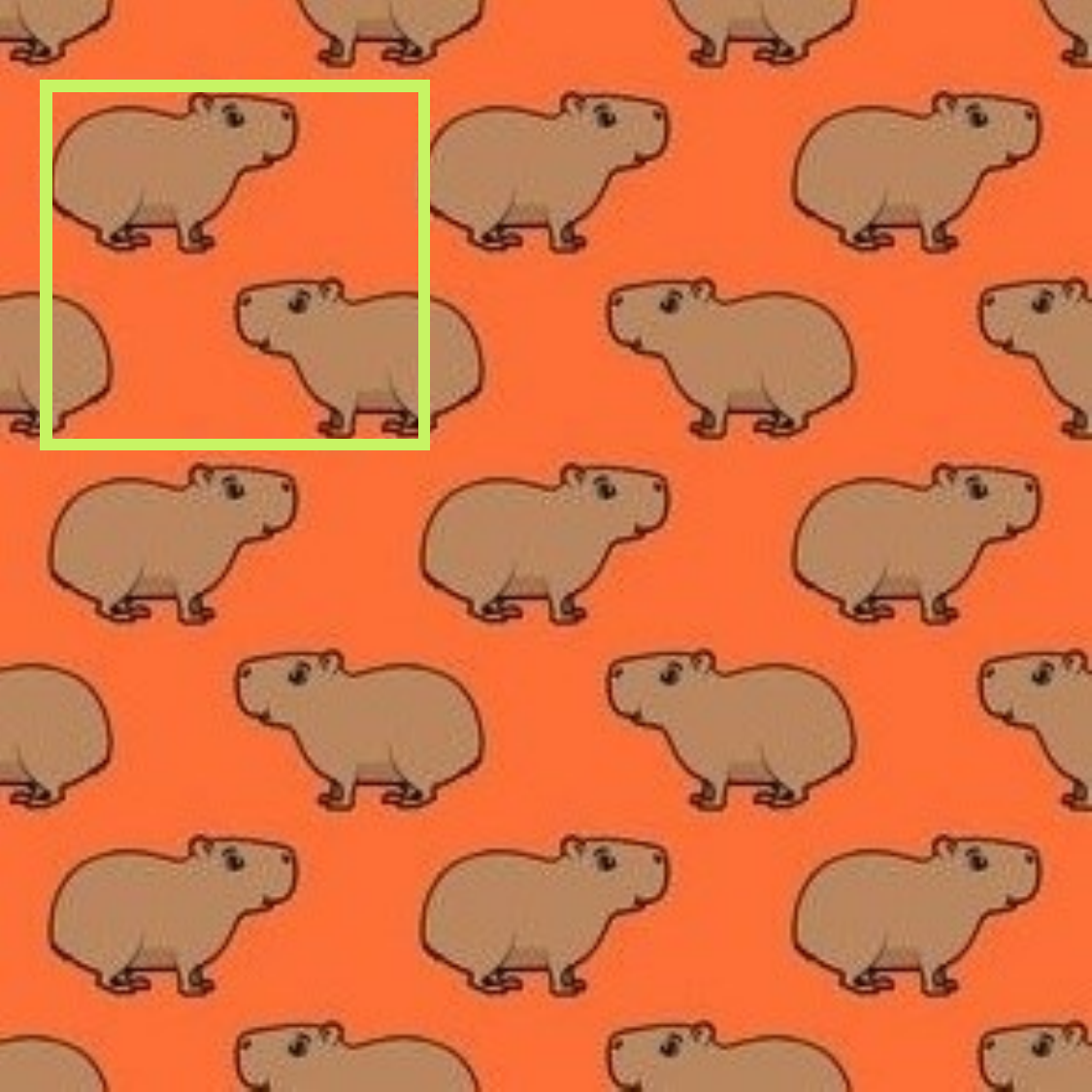} }}%
	\hfil
	\subfloat{\vspace{-3mm}{\includegraphics[width=0.41\columnwidth]{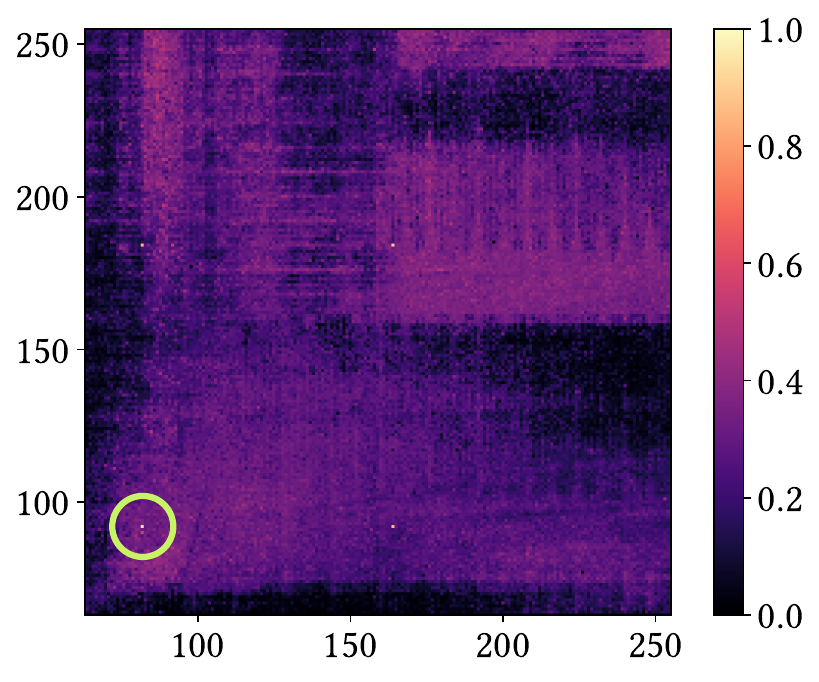} }}%
	\hfil
	\subfloat{{\includegraphics[width=0.275\columnwidth]{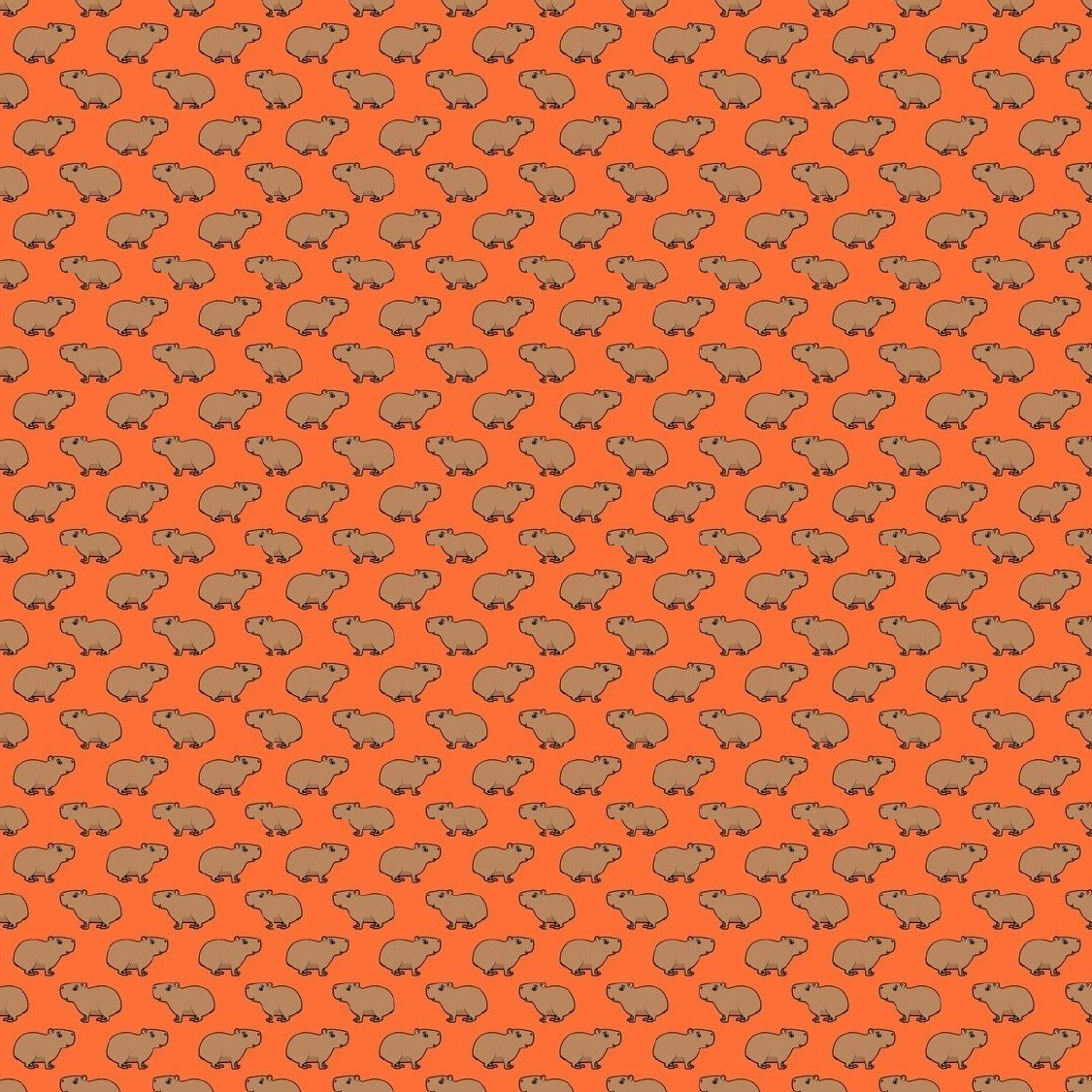} }}%
 
	\caption{On the left, input image. In the middle, TexTile values for different crop sizes. The optimal crop is highlighted on both images with a green inset. On the right, the optimal crop, tiled many times for visualization.}%
	\vspace{-7mm}
	\label{fig:repeating_pattern}%
\end{figure}

We can also leverage TexTile to compute the \textbf{size of the repeating pattern} in images. Given an axis-aligned image $\text{I}$, we can find the size $h,w$ of the repeating pattern in the image by finding the crop that maximizes its tileability: $\argmax_{h,w} \text{TexTile}(\textrm{Crop}(\text{I}, (h,w)))$. We show a result in Figure~\ref{fig:repeating_pattern}, where this algorithm finds the optimal crop size at the minimum repeatable pattern, as well as three lower peaks at different discrete scale factors of this crop size. Note that we limit the crops to $h,w \geq 64$ and perform the crop at the center of the image. Previous work~\cite{lettry2017repeated,rodriguez2019automatic} found these repeating patterns in the activations of pre-trained CNNs. Because internal neural activations operate at lower resolutions that those of the original image, these methods are limited in precision. Our method, in contrast, operates at the resolution of the original image and may thus be more precise. More results on these two applications are present in the supplementary material.

\subsection{Failure Cases}
\begin{figure}[tb!]
\vspace{-5mm}
	\centering
	\includegraphics[width=1.0\columnwidth]{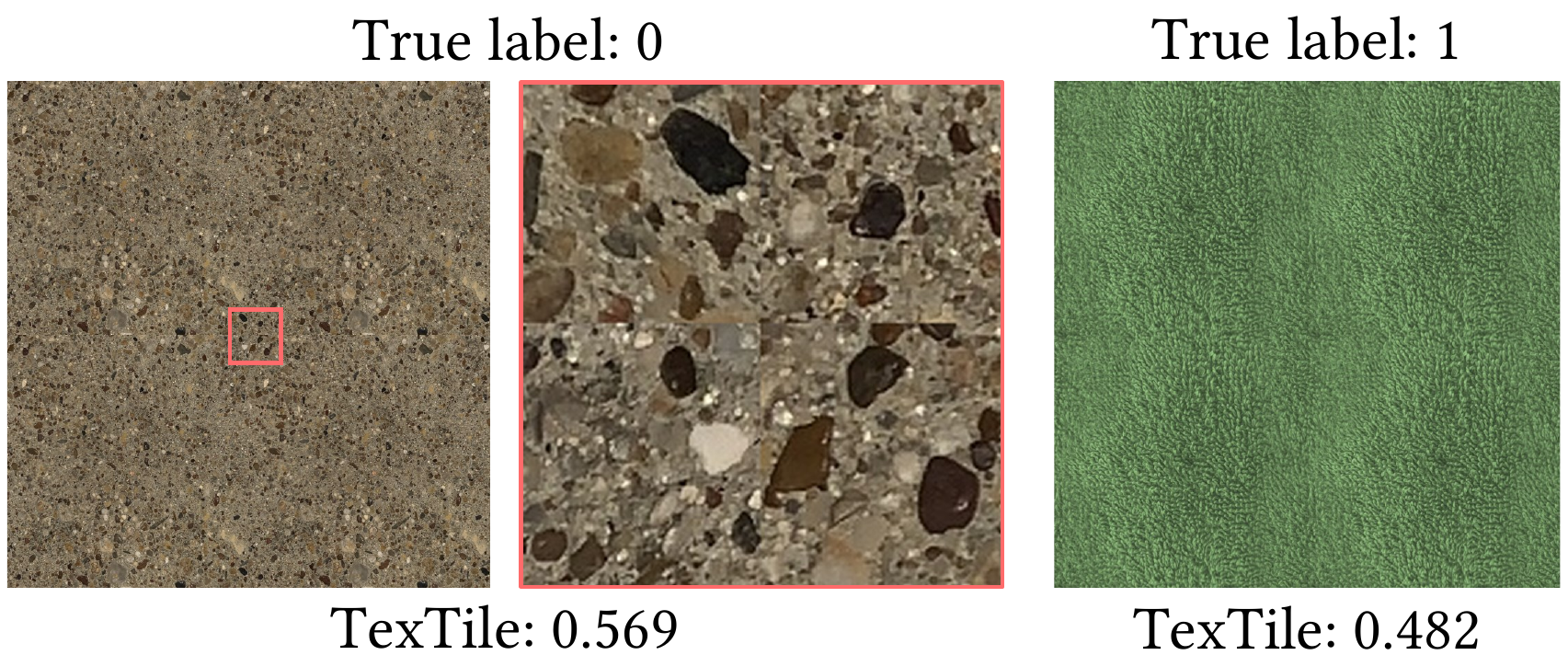}
 \vspace{-7mm}

	\caption{Examples of misclassifications done by our model. On the top, ground truth labels, on the bottom, the predicted TexTile value. For the example on the left, we show an inset of the central crop of the texture, to highlight that it is not seamlessly tileable.  }
	\label{fig:limitations}
	 \vspace{-6mm}

\end{figure}

As shown in Table~\ref{tab:ablation}, our model predictions are accurate, however, some errors occur. We show some examples of misclassified textures in Figure~\ref{fig:limitations}. On the left, we show a texture labeled as non-tileable in our test dataset, that our model predicts as tileable despite it showing discontinuities in the borders. On the right, we show a texture that is labeled as tileable, which our model classifies as non-tileable. Both examples are edge cases and highlight the ambiguity in what constitutes a tileable texture.
Besides, when used as a loss for synthesis models, TexTile cannot compensate for the limitations of the generative backbone. If a model cannot adequately synthesize textures that match the appearance of the input, adding TexTile  helps reduce discontinuities in the borders but will not improve its perceptual quality, as can be seen in the textures generated with~\cite{heitz2021sliced} in Figure~\ref{tab:comparisons}.

\section{Conclusions}\label{sec:conclusions}
We have presented \emph{TexTile}, the first differentiable metric for texture tileability. While it is trained on a simple classification task, we design custom data augmentation, training regimes, and neural architectures, all specifically tailored to accurately measure tileability. We validated our design choices with comprehensive ablation studies, and leveraged saliency maps for model understanding. We showed different applications of our differentiable metric, including benchmarking texture synthesis algorithms, detecting repetitions and misalignment in images, and transforming image generative models into tileable texture synthesis algorithms. We will provide code and model weights.

\paragraph*{Limitations and Future Work} We could extend our work in several ways. While our method accurately measures tileability in textures, it is does not quantify their perceptual quality, limiting its scope. Combining perceptual metrics with TexTile, to measure both perceptual quality and tileability, is an important research avenue for a more integrated analysis of texture quality. %
Besides, our model is pre-trained on a ImageNet, then fine-tuned on a manually-curated dataset, and thus may inherit biases. While we believe the datasets we used were comprehensive, it is possible that some type of texture is underrepresented and the model may not perform accurately. Using synthetic data~\cite{fu2023dreamsim} may help alleviate this issue. Finally, there is no solid understanding of human perception of texture repetitiveness~\cite{sun2021visual}, and, while human perceptual validation is out of the scope of this work, its apparent higher correlation with TexTile, might add new insights to the elements identified so far.

{\scriptsize
\paragraph*{Acknowledgments}~This publication is part of the project TaiLOR, CPP2021-008842 funded by MCIN/AEI/10.13039/501100011033 and the NextGenerationEU / PRTR programs. Elena Garces was partially supported by a Juan de la Cierva - Incorporacion Fellowship (IJC2020-044192-I).
}

{
	\small
	\bibliographystyle{ieeenat_fullname}
	\bibliography{references}
}

\end{document}